\pdfoutput=1
\documentclass[10pt,letterpaper]{article}

\usepackage{arxiv}

\AtBeginDocument{%
  \newgeometry{
    textheight=9in,
    textwidth=516pt,
    top=1in,
    headheight=14pt,
    headsep=25pt,
    footskip=30pt
  }%
}

\usepackage[utf8]{inputenc}
\usepackage[T1]{fontenc}
\usepackage[english]{babel}

\usepackage{amsmath,amssymb,amsfonts}
\usepackage{array}
\usepackage[caption=false]{subfig}
\usepackage{textcomp}
\usepackage{url}
\usepackage{graphicx}
\graphicspath{{media/}}

\usepackage{pgf}

\usepackage{gridfigure}

\usepackage{booktabs}
\usepackage{tabularx}
\newcolumntype{L}{>{\raggedright\arraybackslash}X}
\newcolumntype{C}{>{\centering\arraybackslash}X}
\usepackage{nicefrac}
\usepackage{microtype}
\usepackage{algpseudocode}
\usepackage{algorithm}
\usepackage[switch]{lineno}

\usepackage[hidelinks]{hyperref}

\usepackage{csquotes}

\usepackage[bibstyle=ieee,citestyle=numeric-comp,backend=biber]{biblatex}

\providecommand{\IEEEPARstart}[2]{#1#2}

\makeatletter
\newcommand{\rl@appendixseccntformat}[1]{%
  \csname rl@appendixprefix@#1\endcsname\csname the#1\endcsname\quad
}
\expandafter\newcommand\csname rl@appendixprefix@section\endcsname{\appendixname~}

\long\def\@makecaption#1#2{%
  \vskip\abovecaptionskip
  \footnotesize
  \sbox\@tempboxa{#1: #2}%
  \ifdim\wd\@tempboxa>\hsize
    #1: #2\par
  \else
    \global\@minipagefalse
    \hb@xt@\hsize{\hfil\box\@tempboxa\hfil}%
  \fi
  \vskip\belowcaptionskip}
\makeatother

\newcommand{\rlTableWidth}{0.8\textwidth}
\newcommand{\rlFigureWidth}{0.5\textwidth}

\makeatletter
\renewenvironment{table}[1][\fps@table]
  {\setlength{\abovecaptionskip}{\@belowcaptionskip}%
   \setlength{\belowcaptionskip}{\@abovecaptionskip}%
   \def\@floatboxreset{\reset@font\footnotesize\@setminipage}%
   \@float{table}[#1]%
   \setlength{\columnwidth}{\rlTableWidth}}
  {\end@float}

\renewenvironment{figure}[1][\fps@figure]
  {\@float{figure}[#1]%
   \setlength{\columnwidth}{\rlFigureWidth}}
  {\end@float}
\makeatother

\title{Reinforcement Learning to Choose Optimizers}

\author{%
  Martin van der Schelling\\
  \normalfont Department of Mechanical Engineering\\
  \normalfont Delft University of Technology, Delft, the Netherlands\\
  \normalfont \texttt{M.P.vanderSchelling@tudelft.nl}%
  \And
  Deepesh Toshniwal\\
  \normalfont Institute of Applied Mathematics\\
  \normalfont Delft University of Technology, Delft, the Netherlands\\
  \normalfont \texttt{D.Toshniwal@tudelft.nl}%
  \And
  Miguel A. Bessa\\
  \normalfont School of Engineering\\
  \normalfont Brown University, Providence, RI, USA\\
  \normalfont \texttt{miguel\_bessa@brown.edu}%
}

\renewcommand{\shorttitle}{Reinforcement Learning to Choose Optimizers}

\hypersetup{
	pdftitle={Reinforcement Learning to Choose Optimizers},
	pdfauthor={Martin van der Schelling, Deepesh Toshniwal, Miguel A. Bessa},
	pdfkeywords={Algorithm portfolios, Dynamic algorithm configuration, Reinforcement learning},
}

\begin{document}

\maketitle


\begin{abstract}
No single optimization method is uniformly best for all problems, and the most suitable optimizer choice can change during a run. Existing approaches that change optimizer during execution typically predetermine part of the strategy: the portfolio is restricted to one algorithm class, the switch occurs once at a fixed time, or the frequency of decisions is treated as a hyperparameter rather than a learned one. We introduce ``Reinforcement Learning to Choose Optimizers'', which formulates the optimization algorithm choice as a sequential decision-making problem. At each decision, a recurrent policy reads the current run state and decides both which optimizer should be used next and for how long. The portfolio includes both gradient-based and derivative-free optimizers, and each switch passes on the current best solution and a representative step size. A context proxy conditions a gating network over expert heads, and training employs a decoupled actor–critic whose return is expressed in the same empirical runtime distribution metric used at evaluation. Training tasks and portfolio are designed jointly so that no optimizer dominates. On unseen problems, the learned policy outperforms every portfolio optimizer at all but the smallest budgets, and it remains robust under distribution shift.
\end{abstract}

\keywords{Algorithm portfolios \and Dynamic algorithm configuration \and Reinforcement learning}

\section{Introduction}
\label{sec:introduction}

\IEEEPARstart{O}{ptimization} is central to data-driven design~\autocite{Bessa2017}. Evaluating the objective typically requires a simulation, so the number of loss function evaluations is constrained. The ``No Free Lunch'' theorem states that no single optimization algorithm outperforms all others on every problem~\autocite{Wolpert1997}, making optimizer selection and hyperparameter tuning a design problem in itself. Meta-optimization answers it from several directions: automatic algorithm selection~\autocite{Kerschke2018a}, hyper-heuristics~\autocite{Dokeroglu2024,Misir2021} and learning to optimize~\autocite{Andrychowicz2016,Li2017}. Each of these approaches trains a model to conduct this selection, and that model is static once training has finished, fixing in advance how far a run can adapt. The most effective optimizer is specific to the loss landscape and changes as the search progresses, and a well-placed switch between two algorithms can outperform either of them alone~\autocite{Krasnogor2005,Vermetten2020}.

Existing methods that switch optimizers mid-run always predefine some portion of the strategy and learn only the remaining part. One line of work uses a single offline-computed switch from a global to a local method, applied once per problem~\autocite{Schroder2023}. Another learns repeated switches but restricts the portfolio to one algorithmic family, keeping the optimizer portfolio narrow~\autocite{Guo2024,Zhu2024,Tahernezhad-Javazm2024}. A third line switches at every iteration~\autocite{Getzelman2021}. In all these cases, the interval between decisions is a framework hyperparameter, not controlled by the policy.

Each element of this problem has been solved in isolation. Portfolios combining gradient-based and derivative-free methods have existed for decades, full optimizers have been selected and transferred to new problems, and the refinement effort has been adapted since the earliest memetic algorithms~\autocite{Ong2004,Neri2012,Meunier2021,Guo2024,Krasnogor2005}. Repeatedly deciding which optimizer takes control, and for how long, must be done over a portfolio whose members do not share internal-state representations. Any policy for this must be deployed over a distribution of problems, precisely where learned meta-optimizers are weakest~\autocite{Ma2025b}.

\textit{Contributions}: We introduce ``Reinforcement Learning to Choose Optimizers'' (RL2CO), which formulates the schedule of an optimization run as a sequential control problem. At each decision point, a recurrent policy observes the run state and selects (i) an optimizer from a fixed portfolio and (ii) how many iterations it may run before revisiting the choice. The portfolio contains both gradient-based and derivative-free methods. A switch passes the new optimizer the best solution obtained so far along with a representative step size, allowing it to continue at the scale the search has already achieved rather than reverting to a global scale. Because the same observation can require different actions on different landscapes, the policy is conditioned on a proxy for the problem context. This proxy feeds a gating network over expert heads. We train with a decoupled actor-critic model. The return it maximizes is denominated in the same empirical runtime distribution metric the evaluation reports. Since a portfolio in which one member dominates would leave no benefit for a learned optimizer selector, the training problems and the portfolio are designed jointly, maximizing the margin an informed per-problem choice holds over the portfolio average. This work extends our earlier study~\autocite{vanderSchelling2026}, where a single optimizer was chosen from the problem features and switched only once.

To assess the framework we train a policy across 100 tasks drawn from three Black-Box Optimization Benchmarking (BBOB) suites, over a portfolio and task set selected jointly. It is evaluated on 100 held-out problems, against every member of that portfolio and against a random schedule over the same optimizers. The same policy is tested on problem families absent from training, showing robustness to distribution shift. Our code, dataset, and documentation are available according to FAIR principles (Findable, Accessible, Interoperable, and Reusable)\footnote{\url{https://github.com/bessagroup/rl2co}}.

\section{Related Work}
\label{sec:related-work}

\noindent We consider single-objective optimization problems where an objective function $f(\boldsymbol{x})$, over parameters $\boldsymbol{x} = (x_1, x_2, \dots, x_d)$, is minimized by an iterative procedure. The procedure uses an update rule $g(\boldsymbol{x}, f(\boldsymbol{x}), \boldsymbol{\psi})$ with hyperparameters $\boldsymbol{\psi}$, applied repeatedly to an initial solution $\boldsymbol{x}^0$ until a termination criterion is met. A run may use different update rules over time and is represented by a schedule: a sequence $((g_1, \tau_1), (g_2, \tau_2), \dots)$ where rule $g_i$ is executed for $\tau_i$ iterations before switching. The meta-algorithmic task is to design this schedule.

Optimization problems are often categorized by what information the optimizer can access: in black-box settings only function evaluations are available, while in white-box settings the analytical form is known and gradients can be computed~\autocite{Ma2025}. Because our schedule design allows first-order optimizers, we focus on the latter. Our scope, however, is narrower than full analytical transparency. With automatic differentiation, gradients can be computed for any simulation in a differentiable programming environment~\autocite{Bradbury2018,Blondel2021}, and in data-driven design the objective is often a trained surrogate model that can be differentiable by construction~\autocite{Bessa2017}. We concentrate on differentiable design problems, a class that is both broad and rapidly expanding. The method itself does not rely on this requirement: one could instead use only derivative-free optimizers, which would change the range of applicable problems but not the framework.

\subsection{Choosing an Optimizer Before the Run}
\label{subsec:before-the-run}

\noindent The algorithm selection problem aims to map each problem instance to the optimizer expected to perform best on it~\autocite{Rice1976,Kerschke2018a,Bischl2012}. This requires characterizing optimization problems via features, which in turn rely on function evaluations. Exploratory Landscape Analysis (ELA) computes such features by sampling points from the design space~\autocite{Mersmann2011}. However, due to the ``curse of dimensionality'', this sampling overhead can grow rapidly, even exceeding the cost of directly solving the optimization problem~\autocite{He2007}, and the resulting features are highly sensitive to the sampling strategy~\autocite{Renau2020}.

Both limitations have been mitigated. Trajectory-based methods derive features from the optimizer's search path instead of a separate exploratory phase~\autocite{Jankovic2021,Renau2024}, and learned descriptors replace hand-crafted features with end-to-end trained representations~\autocite{vanStein2023,Alissa2019}.

\subsection{Changing Optimizer During the Run}
\label{subsec:during-the-run}

\noindent Although advances in automatic algorithm selection reduce the effort of problem characterization, these models still commit to a single optimizer for the entire run. Prior work shows that the best local search operator is both instance- and time-dependent~\autocite{Krasnogor2005}, and that carefully scheduling switches between two algorithms can outperform using either alone~\autocite{Vermetten2020}, even though theoretical guidelines for when and how to switch remain largely absent~\autocite{Antipov2026}. The most suitable optimizer depends on both the problem and the current search stage.

Meta-Black-Box Optimization (MetaBBO) models this as a bi-level problem where a meta-policy designs a low-level optimizer~\autocite{Ma2025}. Hyper-heuristics share the bi-level structure~\autocite{Burke2010,Dokeroglu2024,Misir2021} but differ in scope and adaptivity: they select or generate heuristics mainly for combinatorial problems and operate offline in a fixed domain, rather than adapting online across problem classes~\autocite{Ma2025}. The same structure is formalized as dynamic algorithm configuration, in which a policy, conditioned on an algorithm's state, sets its hyperparameters and is trained over a problem distribution~\autocite{Biedenkapp2020,Adriaensen2022}. A more radical variant removes the low-level optimizer and lets the meta-policy propose candidate solutions directly~\autocite{Andrychowicz2016,Li2017,Ma2025}. These ``learning to optimize'' (L2O) methods can express strategies beyond any fixed update rule, but the vast space of update rules makes them hard to train and unreliable out of distribution~\autocite{Metz2019,Harrison2022}. Adaptively selecting from a small set of established optimizers instead preserves convergence guarantees. This introduces three questions absent for static choices, posed explicitly by \textcite{Schroder2023}: \textit{to which} optimizer to switch, \textit{when} to switch, and \textit{how} to switch.

\paragraph{Which optimizer to switch to}

The optimizer switched to is chosen from a predefined portfolio, and how that portfolio is built limits the potential benefits of switching~\autocite{Vermetten2020}. When selection \textit{is} performed during the run, it is usually among very similar algorithms~\autocite{Guo2024,Zhu2024,Tahernezhad-Javazm2024}. Each of these portfolios is restricted to a single evolutionary family and excludes gradient-based methods. When a portfolio does span multiple families, its composition is fixed in advance rather than adapted during the run~\autocite{Meunier2021}.

Memetic algorithms combine population-based global search with a local refinement operator, and the survey of \textcite{Neri2012} sets out the design choices this involves. \textcite{Ong2004} proposed a meta-learning approach that selects the local search method for each candidate at runtime. The computational effort for local search, termed \textit{intensity}, concerns how long improvement is attempted and how often the procedure is invoked~\autocite{Krasnogor2005}; these settings are usually fixed offline and only rarely adapted. A third option is to resume the procedure from its previous state, as in local search chains~\autocite{Molina2010}. Two aspects distinguish this from the schedule considered here: the refinement operator is chosen within a fixed evolutionary outer loop rather than as the main optimizer for the entire run~\autocite{Cao2026}; and adaptation occurs within a single run and is discarded for subsequent problems, with no learning transfer over a problem distribution~\autocite{Samma2016}.

\paragraph{When to switch}

Two lines of work address when to switch without controlling it. Trajectory-based selection predicts whether a switch at a given point would help, using features from a sliding search window~\autocite{Vermetten2023,Kostovska2022,Jankovic2022}. The window length, the successor's budget, and the number of switches are fixed before the run. Allocation, instead, is modeled as a bandit problem: \textcite{Gagliolo2007} run all candidates in parallel, learn their runtime distributions, and allocate budget to those performing well.

\paragraph{How to switch}

Whether switching optimizers pays off depends mainly on how much knowledge is transferred between them. \textcite{Schroder2023} show that the benefit comes from warm-starting the new optimizer from the old one's state; without this ``synergy'', the second optimizer uses little more than the current iterate and discards progress stored in the parameters. This transfer is designed in advance, however: it targets a specific optimizer pair and a single switch, with a warm start crafted for that pair and stage, and nothing chosen dynamically during the run. \textcite{Schroder2023} flag as open the challenge of deciding the switch per run, using information from the ongoing trajectory. Such hand-built handovers are not unique to black-box optimization. In a gradient-based setting, \textcite{Keskar2017} switch once from Adam to stochastic gradient descent (SGD) and convert Adam's moment estimates into an SGD step size via a closed-form rule tailored to that pair. Even when a policy repeatedly chooses optimizers, the transfer is still fixed: \textcite{Getzelman2021} pick among Adam, gradient descent, and random search at each iteration, but reconcile their incompatible states using a predetermined rule. In all these cases, authors decide what is transferred; only which optimizer runs next is left to the run.

Each of these approaches answers one of the three questions and fixes the others. None treats them as a single runtime decision: which optimizer to switch to, how long it keeps the budget before revisiting the choice, and what state it inherits. The duration is typically computed offline or reduced to a single iteration, so no method learns how long to commit. Even when the decision is repeated, the interval between decisions is a framework hyperparameter. Deciding too often breaks the optimizer's integrity; deciding too rarely sacrifices the flexibility that motivates switching in the first place~\autocite{Guo2024}.

\subsection{Robustness of Learned Policies Under Distribution Shift}
\label{subsec:robustness}

\noindent \textcite{Ma2025b} evaluate several learned meta-optimization methods on test suites from distributions different from those used in training and find that learned policies fail under this distribution shift, whereas classical algorithms remain robust. Meta-learning is most vulnerable precisely when the underlying problem changes. Learned selectors show the same issue: a per-run selector trained on the BBOB suite~\autocite{Hansen2009} did not transfer to the Nevergrad collection~\autocite{Rapin2018}, which its authors attribute to the low similarity between the two~\autocite{Kostovska2022}.

\textcite{Wang2026a} attribute overfitting to the limited diversity of standard benchmarks and construct a richer training set by embedding problem features in a latent space and reverse-engineering function formulas for sampled points. An alternative strategy avoids any fixed training distribution and instead adapts online to the specific target task~\autocite{Wang2026}.

Although broadening the training distribution improves generalization, it does not explain how the policy represents what it learns. In multi-task reinforcement learning (RL), policies use mixtures of state encoders, routed by a gate conditioned on task metadata~\autocite{Sodhani2021}. This separates two roles often conflated: the mixture prevents interference between task-specific representations, and the metadata lets a policy act in novel environments. The closest meta-optimization approach trains a single policy over a modular algorithm design space via multi-task RL~\autocite{Guo2025}, adopting the training setup but not the conditioning: one encoder is shared across tasks, and the policy receives no task description. Neither mechanism has been used for robustness in algorithm selection, where the problem is specified before search begins and the trajectory supplies corrective feedback during optimization.

\subsection{Knowledge Gap: Constructing the Schedule}
\label{subsec:knowledge-gap}

\noindent The literature provides all the elements of this problem but never unifies them. Hybrid compositions of gradient-based and derivative-free methods have existed for decades, yet remain manually designed~\autocite{Neri2012,Ong2004,Meunier2021}. Learning to select among full optimizers yields transferable strategies, but only within portfolios from a single algorithmic family~\autocite{Guo2024,Zhu2024,Tahernezhad-Javazm2024}. Switching between optimizer families has been shown, with the new optimizer warm-started, but the switch happens only once, at a fixed time, using manually engineered knowledge transfer~\autocite{Schroder2023,Kostovska2022}. In works with multiple switches, the decision is instead made at every iteration~\autocite{Getzelman2021}.

The field lacks a method that treats the schedule of an optimization run as a learned, on-the-fly decision. In particular, it lacks a system that:

\begin{enumerate}
    \item reassesses the optimizer \textit{continuously} as the search unfolds, rather than only once before or during the run, unlike dynamic selection, which is limited to a single algorithm family;
    \item operates over a portfolio spanning both gradient-based and derivative-free methods, repeatedly crossing this boundary in both directions while carrying the progress of the run across each switch, rather than a warm start designed for a single ordered pair (\textit{heterogeneous handover});
    \item learns \textit{how long} each optimizer should run before revisiting the choice, allowing budget trade-offs between families with iteration costs differing by orders of magnitude, instead of using a fixed, pre-set schedule;
    \item maintains robust performance on problem suites outside the training distribution.
\end{enumerate}

That this combination is a deliberate target rather than an incidental omission is clear from the field's own agenda. Extending their syntactic model of memetic algorithms, \textcite{Krasnogor2005} introduce a \textit{metascheduler}: a component that uses information from previous populations to choose which operators to apply, with parameters that ``may now represent complex data structures rather than simple probability distributions''~\autocite[p.~482]{Krasnogor2005}. They report no prior instances of such algorithms in the literature. A policy conditioned on run history and carrying a learned recurrent state is one such instance. \textcite{Ma2025} call for a higher-level framework defining meta-level tasks across multiple optimizers, enabling a single policy to be trained over, and generalize across, a shared design space. When the elements of this space are full optimizers, the schedule is precisely that design space.

\section{Methods}
\label{sec:methods}

\begin{figure}[!t]
    \centering
    \includegraphics[width=\columnwidth]{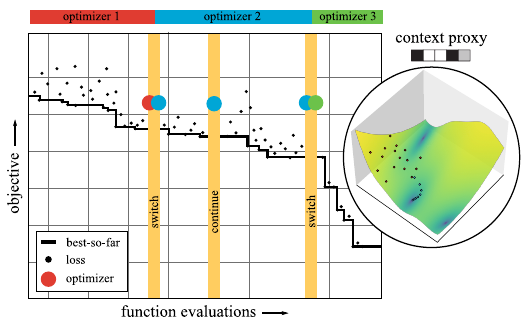}
    \caption{Overview of RL2CO on a single optimization run. The colored bar records the schedule: which optimizer iterates over which part of the total budget. At the decision point, the policy selects the next optimizer $k_t$ and the number of iterations $\tau_t$ it may run. A decision either switches or continues with the current optimizer. The loss landscape below is summarized by the context proxy $\tilde{c}$ that conditions the policy.}
    \label{fig:overview}
\end{figure}

\subsection{Optimizer Selection as a Contextual Markov Decision Process}
\label{subsec:cmdp}

\noindent Section~\ref{sec:related-work} defined an optimization run as a schedule where update rule $g_i$ is applied for $\tau_i$ iterations. We construct such a schedule with an agent interacting with an environment over discrete time steps~\autocite{Sutton2018}. Figure~\ref{fig:overview} shows that construction over a single run, from the decision points to the schedule they produce.

Most learned optimizers are trained on a single task or closely related tasks~\autocite{Li2017,Andrychowicz2016}. We instead aim for a model that generalizes across many problems, so the environment is a family of decision processes. Each task has a context $c \in \mathcal{C}$, which determines the objective $f(\boldsymbol{x})$ to minimize over parameters $\boldsymbol{x} \in \mathbb{R}^d$ and with it the agent's dynamics. At each time step $t$, the agent observes $o_t \in \Omega$, a projection of the environment state $s_t$, takes an action $a_t$ under a stochastic policy $\pi(a \mid o, c)$, and receives a scalar reward $r_t$. This yields a \textit{contextual} Markov decision process $\langle \mathcal{C}, \mathcal{S}, \mathcal{A}, \Omega, T_c, R_c, \gamma \rangle$~\autocite{Hallak2015,Biedenkapp2020}, where $\mathcal{S}$, $\mathcal{A}$, and $\Omega$ are the state, action, and observation spaces, $T_c(s' \mid s, a)$ is the transition function, $R_c(s, a)$ is the reward function, and $\gamma \in (0, 1]$ is the discount factor. The goal is to maximize the expected discounted return $J(\pi)$ over the context distribution $q(c)$, estimated during training from the available problems.

\begin{equation}
    J(\pi) = \mathbb{E}_{c \sim q(c)}\left[\mathbb{E}\left[\sum\nolimits_{t=0}^{H}
    \gamma^{\sum\nolimits_{i<t} \Delta_i} R_c(s_t, a_t)\right]\right],
    \label{eq:rl-objective-contextual}
\end{equation}

The transition function $T_c(s' \mid s, a)$ is realized by the optimization run itself and is context-dependent: an update rule $g(\boldsymbol{x}, f(\boldsymbol{x}), \boldsymbol{\psi})$ advances the run by a chosen number of iterations, moving the state from $s$ to $s'$. The agent does not output candidate solutions directly. Instead, each action $a_t = (k_t, \tau_t) \in \mathcal{A}$ comprises two independent choices: $k_t \in \{1, \dots, K\}$ selects which optimizer runs next, and $\tau_t \in \mathcal{T}$, a small fixed set of durations, sets how many iterations it may perform before the next policy query. Sampling $\tau_t$ from such a set simplifies the policy, while including values over several orders of magnitude preserves the distinction between brief exploration and long commitment. A decision consumes $\Delta_t$ iterations. One decision performs many optimizer iterations, so the discount accumulates over iterations rather than over decisions. If $k_{t-1} \ne k_t$, we switch optimizers, initializing the new one from the previous optimizer's state; see Appendix~\ref{app:handshake} for details.

The state $s_t \in \mathcal{S}$ has two parts. The first is the runtime state of each portfolio optimizer: its candidate solutions and internal parameters. The second tracks global progress: observed objective values, the number of iterations, and function evaluations used, and the index of the optimizer that produced the latest iteration. The objective is to achieve a low function value with few function evaluations. These two aims are measured jointly by the area under the convergence curve on a logarithmic evaluation axis~\autocite{Hansen2022}. The reward $R_c(s, a)$ is the drop in a potential that prices a state by the area already accrued and the area its current level would still accrue over the budget remaining, normalized for comparability across contexts. Appendix~\ref{app:environment} defines it, together with the rest of the environment configuration.

\subsection{Conditioning the Policy on Context}
\label{subsec:conditioning-on-context}

\begin{gridfigure}[cols=2,sharey=false,gutter=1pc,asfigures=true]
  \gfitem[graphic]{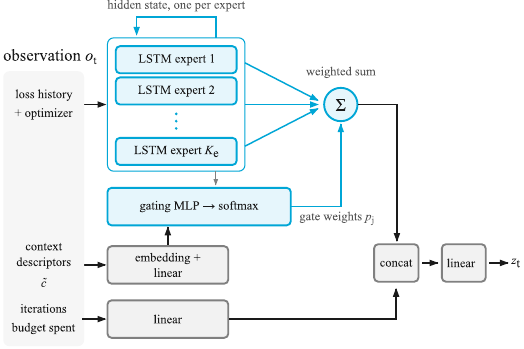}{The conditioning architecture. Each of $K_{\text{e}}$ recurrent \textit{experts} encodes the trajectory features, and a \textit{gating network} reads the context descriptors $\tilde{c}$ together with the expert outputs to weight the experts by $p_j$. The weighted sum is concatenated with the context and progress embeddings, then projected to the latent $z_t$.}{fig:conditioning_architecture}
  \gfitem[graphic]{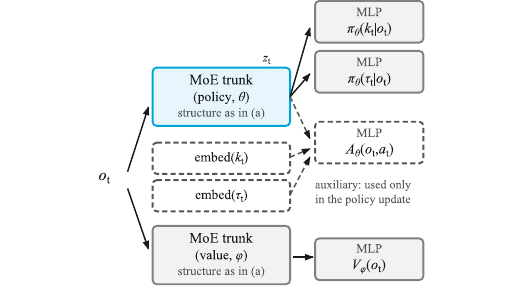}{The actor and critic networks. The policy and value networks are independent copies of the trunk in Fig.~\ref{fig:conditioning_architecture} and share no parameters. The policy weights $\boldsymbol{\theta}$ carry the optimizer head, the duration head, and the auxiliary advantage head $A_{\boldsymbol{\theta}}(o_t, a_t)$, which is evaluated only in the policy update, while the value weights $\boldsymbol{\phi}$ carry the critic $V_{\boldsymbol{\phi}}(o_t)$.}{fig:actor_critic}
\end{gridfigure}

\noindent The policy for a run is largely determined by the objective function $f(\boldsymbol{x})$: irregular, high-dimensional, or noisy landscapes require different update rules and converge differently from smooth, low-dimensional convex ones. The context $c$ is informative yet not directly observable, because $f(\boldsymbol{x})$ is only revealed through interaction. Instead, we condition on a proxy $\tilde{c}$ that encodes what is known before the run. We use two information sources: (i) descriptors of the loss landscape, such as dimensionality or noise level, and (ii) user-provided information, such as physical system properties or known analytical features of the objective. Since both are available during training and deployment, $\tilde{c}$ serves as a stand-in for $c$ throughout.

Conditioning on context is essential because the same observation can demand opposite behaviors. An oscillating objective trajectory is expected on a stochastic landscape and requires no intervention, but the same pattern on a smooth convex problem indicates that the update rule is diverging and must be changed. A policy that ignores $\tilde{c}$ averages across these regimes and will misinterpret them. Rather than rescaling a single shared policy, $\tilde{c}$ drives a gating network that combines multiple expert heads, each specializing in a region of context space~\autocite{Shazeer2017}, as drawn in Fig.~\ref{fig:conditioning_architecture}.

\subsection{Training With a Decoupled Actor-Critic}
\label{subsec:training}

\noindent The policy is optimized with a policy-gradient method, training a neural network from the gradient of the expected return~\autocite{Sutton2018}. Because return from a state depends strongly on how far the run has progressed and how much budget remains, a Monte Carlo estimate conflates action quality with the run stage. To reduce this variance, we use an actor-critic approach: a separate critic network estimates each state's value, and subtracting this estimate yields an advantage signal for the actor network, which is used to compare the available actions.

Typical actor-critic architectures share a single learned representation between actor and critic. Yet value estimation and action selection have different objectives: predicting return needs fine-grained, instance-specific environment details, and a policy sharing a representation that encodes them overfits and fails to generalize to new tasks~\autocite{Raileanu2021}. Because we study this generalization problem, we adopt decoupled advantage actor-critic (DAAC), which trains separate networks for policy and value and adds an auxiliary advantage head to the policy network~\autocite{Raileanu2021}. Figure~\ref{fig:actor_critic} shows both networks and the heads each one carries. Advantages are computed with generalized advantage estimation (GAE)~\autocite{Schulman2018}, and the update uses proximal policy optimization (PPO)~\autocite{Schulman2017}, enabling multiple epochs of minibatch updates per batch of episodes. Further details of the training procedure are provided in Algorithm~\ref{alg:training} in Appendix~\ref{app:training}.

\subsection{Computational Overhead}
\label{subsec:overhead}

\noindent Training the policy is a one-time offline cost. Once trained, deployment requires only a single forward pass through the policy network per decision, taking milliseconds on a typical CPU. Since the policy acts only at switching points, the number of passes per run is far below the iteration budget, so the dominant cost remains the lower-level optimizer update and objective evaluation.

\section{Results}
\label{sec:results}

\subsection{Constructing the Task Set and Optimizer Portfolio}
\label{subsec:taskset}

\noindent A learned policy is only as informative as its training problems and as effective as its optimizer portfolio. To generalize to unseen loss landscapes, training tasks must cover diverse characteristics. Together, tasks and portfolio should reflect the ``No Free Lunch'' theorem: no optimizer should dominate across all tasks, or selection would become trivial~\autocite{Wolpert1997}. We design the training set and optimizer portfolio jointly and first describe the result of this joint construction before analyzing the policy's behavior on it.

To obtain problems reflecting diverse landscape characteristics, we use the BBOB suite~\autocite{Hansen2009}, which provides analytical functions varying in conditioning, modality, and separability. To study stochastic effects, we also include the noisy BBOB suite~\autocite{Hansen2009a}.

We further aim to capture structural properties of over-parameterized neural network loss landscapes. These are high-dimensional but effectively low-rank: gradients concentrate in a low-dimensional subspace~\autocite{Gur-Ari2018}, and objectives can be optimized within randomly chosen subspaces much smaller than the ambient dimension~\autocite{Li2018}. Simply increasing the dimension of standard BBOB functions does not reproduce this, since each added dimension carries meaningful signal. We enforce this structure explicitly instead, by embedding a low-dimensional BBOB function into a high-dimensional search space. Appendix~\ref{app:benchmark-functions} gives full details of the benchmark suites.

We evaluate the three benchmark suites with 52 gradient-based and derivative-free optimizers plus a random-search baseline. Performance is measured by the number of function evaluations needed to reach a specified target objective, rather than by the objective value alone~\autocite{More2010}. The empirical runtime distribution (ERTD) aggregates this over independent runs and multiple targets, including both successful and unsuccessful trials~\autocite{Hansen2022}. Unsuccessful trials are accounted for by simulated restarts, so the budget axis can exceed the maximum evaluations of a single run.

Comparing the ERTD of the \textit{virtual best solver} (VBS), which selects the best optimizer per task and realization, with that of the \emph{single best solver} (SBS), the one optimizer with the best overall performance per task, quantifies the potential of algorithm-selection strategies~\autocite{Bischl2012}. It also measures how strongly the task set exhibits the conditions of the ``No Free Lunch'' theorem. We use this VBS--SBS gap as the design criterion for both the training task set and the optimizer portfolio. We search the profiled functions and optimizers for a subset that maximizes this gap, while retaining exactly 100 functions and 4 optimizers, and construct a held-out test set from functions not in the training set.

The resulting portfolio contains the following four optimizers:

\begin{itemize}
    \item \textbf{L-BFGS}: a quasi-Newton method with a limited-memory inverse-Hessian approximation and line search along the resulting direction~\autocite{Liu1989}.
    \item \textbf{Rprop}: a first-order method that ignores gradient magnitude and adapts per-coordinate step sizes from the sign of successive gradients~\autocite{Riedmiller1993}.
    \item \textbf{CR-FM-NES}: a natural evolution strategy for high-dimensional problems that samples mirrored pairs around an adaptive mean with a rank-one covariance factor~\autocite{Nomura2022}.
    \item \textbf{MR15-GA}: a genetic algorithm with mutation rate adapted by the one-fifth success rule~\autocite{Rechenberg1989}.
\end{itemize}

Appendix~\ref{app:selection-procedure} details the resulting selection and search procedure.

Figure~\ref{fig:rtd_bbob_headroom_train_random} shows the ERTDs of the training set. The four selected members separate across the budget range without any one of them leading throughout, while the per-task best lies well above all of them.

\begin{gridfigure}[cols=2,sharey=false,gutter=1pc,asfigures=true,placement=!t]
  \begingroup%
\makeatletter%
\begin{pgfpicture}%
\pgfpathrectangle{\pgfpointorigin}{\pgfqpoint{6.997095in}{0.140000in}}%
\pgfusepath{use as bounding box, clip}%
\begin{pgfscope}%
\pgfsetbuttcap%
\pgfsetroundjoin%
\pgfsetlinewidth{1.204500pt}%
\definecolor{currentstroke}{rgb}{0.039216,0.627451,0.823529}%
\pgfsetstrokecolor{currentstroke}%
\pgfsetdash{}{0pt}%
\pgfpathmoveto{\pgfqpoint{0.000000in}{0.060000in}}%
\pgfpathlineto{\pgfqpoint{0.200000in}{0.060000in}}%
\pgfusepath{stroke}%
\end{pgfscope}%
\begin{pgfscope}%
\definecolor{textcolor}{rgb}{0.000000,0.000000,0.000000}%
\pgfsetstrokecolor{textcolor}%
\pgfsetfillcolor{textcolor}%
\pgftext[x=0.245000in,y=0.035000in,left,base]{\color{textcolor}{\rmfamily\fontsize{8.000000}{9.600000}\selectfont L-BFGS}}%
\end{pgfscope}%
\begin{pgfscope}%
\pgfsetbuttcap%
\pgfsetroundjoin%
\pgfsetlinewidth{1.204500pt}%
\definecolor{currentstroke}{rgb}{0.368627,0.337255,0.870588}%
\pgfsetstrokecolor{currentstroke}%
\pgfsetdash{}{0pt}%
\pgfpathmoveto{\pgfqpoint{0.806557in}{0.060000in}}%
\pgfpathlineto{\pgfqpoint{1.006557in}{0.060000in}}%
\pgfusepath{stroke}%
\end{pgfscope}%
\begin{pgfscope}%
\definecolor{textcolor}{rgb}{0.000000,0.000000,0.000000}%
\pgfsetstrokecolor{textcolor}%
\pgfsetfillcolor{textcolor}%
\pgftext[x=1.051557in,y=0.035000in,left,base]{\color{textcolor}{\rmfamily\fontsize{8.000000}{9.600000}\selectfont Rprop}}%
\end{pgfscope}%
\begin{pgfscope}%
\pgfsetbuttcap%
\pgfsetroundjoin%
\pgfsetlinewidth{1.204500pt}%
\definecolor{currentstroke}{rgb}{0.800000,0.447059,0.000000}%
\pgfsetstrokecolor{currentstroke}%
\pgfsetdash{}{0pt}%
\pgfpathmoveto{\pgfqpoint{1.508252in}{0.060000in}}%
\pgfpathlineto{\pgfqpoint{1.708252in}{0.060000in}}%
\pgfusepath{stroke}%
\end{pgfscope}%
\begin{pgfscope}%
\definecolor{textcolor}{rgb}{0.000000,0.000000,0.000000}%
\pgfsetstrokecolor{textcolor}%
\pgfsetfillcolor{textcolor}%
\pgftext[x=1.753252in,y=0.035000in,left,base]{\color{textcolor}{\rmfamily\fontsize{8.000000}{9.600000}\selectfont CR-FM-NES}}%
\end{pgfscope}%
\begin{pgfscope}%
\pgfsetbuttcap%
\pgfsetroundjoin%
\pgfsetlinewidth{1.204500pt}%
\definecolor{currentstroke}{rgb}{0.298039,0.352941,0.015686}%
\pgfsetstrokecolor{currentstroke}%
\pgfsetdash{}{0pt}%
\pgfpathmoveto{\pgfqpoint{2.515486in}{0.060000in}}%
\pgfpathlineto{\pgfqpoint{2.715486in}{0.060000in}}%
\pgfusepath{stroke}%
\end{pgfscope}%
\begin{pgfscope}%
\definecolor{textcolor}{rgb}{0.000000,0.000000,0.000000}%
\pgfsetstrokecolor{textcolor}%
\pgfsetfillcolor{textcolor}%
\pgftext[x=2.760486in,y=0.035000in,left,base]{\color{textcolor}{\rmfamily\fontsize{8.000000}{9.600000}\selectfont MR15-GA}}%
\end{pgfscope}%
\begin{pgfscope}%
\pgfsetbuttcap%
\pgfsetroundjoin%
\pgfsetlinewidth{1.204500pt}%
\definecolor{currentstroke}{rgb}{0.000000,0.000000,0.000000}%
\pgfsetstrokecolor{currentstroke}%
\pgfsetdash{}{0pt}%
\pgfpathmoveto{\pgfqpoint{3.422078in}{0.060000in}}%
\pgfpathlineto{\pgfqpoint{3.622078in}{0.060000in}}%
\pgfusepath{stroke}%
\end{pgfscope}%
\begin{pgfscope}%
\definecolor{textcolor}{rgb}{0.000000,0.000000,0.000000}%
\pgfsetstrokecolor{textcolor}%
\pgfsetfillcolor{textcolor}%
\pgftext[x=3.667078in,y=0.035000in,left,base]{\color{textcolor}{\rmfamily\fontsize{8.000000}{9.600000}\selectfont RL2CO}}%
\end{pgfscope}%
\begin{pgfscope}%
\pgfsetbuttcap%
\pgfsetroundjoin%
\pgfsetlinewidth{1.204500pt}%
\definecolor{currentstroke}{rgb}{0.694118,0.125490,0.513725}%
\pgfsetstrokecolor{currentstroke}%
\pgfsetdash{{4.440000pt}{1.920000pt}}{0pt}%
\pgfpathmoveto{\pgfqpoint{4.197211in}{0.060000in}}%
\pgfpathlineto{\pgfqpoint{4.397211in}{0.060000in}}%
\pgfusepath{stroke}%
\end{pgfscope}%
\begin{pgfscope}%
\definecolor{textcolor}{rgb}{0.000000,0.000000,0.000000}%
\pgfsetstrokecolor{textcolor}%
\pgfsetfillcolor{textcolor}%
\pgftext[x=4.442211in,y=0.035000in,left,base]{\color{textcolor}{\rmfamily\fontsize{8.000000}{9.600000}\selectfont Random schedule}}%
\end{pgfscope}%
\begin{pgfscope}%
\pgfsetbuttcap%
\pgfsetroundjoin%
\pgfsetlinewidth{1.204500pt}%
\definecolor{currentstroke}{rgb}{0.533333,0.533333,0.533333}%
\pgfsetstrokecolor{currentstroke}%
\pgfsetdash{{1.200000pt}{1.980000pt}}{0pt}%
\pgfpathmoveto{\pgfqpoint{5.414045in}{0.060000in}}%
\pgfpathlineto{\pgfqpoint{5.614045in}{0.060000in}}%
\pgfusepath{stroke}%
\end{pgfscope}%
\begin{pgfscope}%
\definecolor{textcolor}{rgb}{0.000000,0.000000,0.000000}%
\pgfsetstrokecolor{textcolor}%
\pgfsetfillcolor{textcolor}%
\pgftext[x=5.659045in,y=0.035000in,left,base]{\color{textcolor}{\rmfamily\fontsize{8.000000}{9.600000}\selectfont Random search}}%
\end{pgfscope}%
\begin{pgfscope}%
\pgfsetbuttcap%
\pgfsetroundjoin%
\pgfsetlinewidth{1.204500pt}%
\definecolor{currentstroke}{rgb}{0.000000,0.000000,0.000000}%
\pgfsetstrokecolor{currentstroke}%
\pgfsetdash{{1.200000pt}{1.980000pt}}{0pt}%
\pgfpathmoveto{\pgfqpoint{6.544977in}{0.060000in}}%
\pgfpathlineto{\pgfqpoint{6.744977in}{0.060000in}}%
\pgfusepath{stroke}%
\end{pgfscope}%
\begin{pgfscope}%
\definecolor{textcolor}{rgb}{0.000000,0.000000,0.000000}%
\pgfsetstrokecolor{textcolor}%
\pgfsetfillcolor{textcolor}%
\pgftext[x=6.789977in,y=0.035000in,left,base]{\color{textcolor}{\rmfamily\fontsize{8.000000}{9.600000}\selectfont VBS}}%
\end{pgfscope}%
\end{pgfpicture}%
\makeatother%
\endgroup%
\\[3pt]
  \gfitem{media/ertd/rtd_bbob_headroom_train_random.pgf}{Empirical runtime distributions over the selected training problems, for the four portfolio members, random search, the virtual best solver (VBS), and three instances of the random schedule, each holding its randomly drawn optimizer for a fixed commitment length $\tau$.}{fig:rtd_bbob_headroom_train_random}
  \gfitem{media/ertd/rtd_bbob_headroom_test.pgf}{Empirical runtime distributions on the held-out test set, for the four portfolio members, the learned policy, a random schedule over the same optimizer menu with $\tau \in \{10, 100, 1000\}$, random search, and the VBS.}{fig:rtd_bbob_headroom_test}
\end{gridfigure}

\begin{gridfigure}[cols=3,sharey=false,gutter=0.06in,valign=b,placement=!t]
  \begingroup%
\makeatletter%
\begin{pgfpicture}%
\pgfpathrectangle{\pgfpointorigin}{\pgfqpoint{6.997095in}{0.140000in}}%
\pgfusepath{use as bounding box, clip}%
\begin{pgfscope}%
\pgfsetbuttcap%
\pgfsetroundjoin%
\pgfsetlinewidth{1.204500pt}%
\definecolor{currentstroke}{rgb}{0.039216,0.627451,0.823529}%
\pgfsetstrokecolor{currentstroke}%
\pgfsetdash{}{0pt}%
\pgfpathmoveto{\pgfqpoint{0.000000in}{0.060000in}}%
\pgfpathlineto{\pgfqpoint{0.200000in}{0.060000in}}%
\pgfusepath{stroke}%
\end{pgfscope}%
\begin{pgfscope}%
\definecolor{textcolor}{rgb}{0.000000,0.000000,0.000000}%
\pgfsetstrokecolor{textcolor}%
\pgfsetfillcolor{textcolor}%
\pgftext[x=0.245000in,y=0.035000in,left,base]{\color{textcolor}{\rmfamily\fontsize{8.000000}{9.600000}\selectfont L-BFGS}}%
\end{pgfscope}%
\begin{pgfscope}%
\pgfsetbuttcap%
\pgfsetroundjoin%
\pgfsetlinewidth{1.204500pt}%
\definecolor{currentstroke}{rgb}{0.368627,0.337255,0.870588}%
\pgfsetstrokecolor{currentstroke}%
\pgfsetdash{}{0pt}%
\pgfpathmoveto{\pgfqpoint{0.806557in}{0.060000in}}%
\pgfpathlineto{\pgfqpoint{1.006557in}{0.060000in}}%
\pgfusepath{stroke}%
\end{pgfscope}%
\begin{pgfscope}%
\definecolor{textcolor}{rgb}{0.000000,0.000000,0.000000}%
\pgfsetstrokecolor{textcolor}%
\pgfsetfillcolor{textcolor}%
\pgftext[x=1.051557in,y=0.035000in,left,base]{\color{textcolor}{\rmfamily\fontsize{8.000000}{9.600000}\selectfont Rprop}}%
\end{pgfscope}%
\begin{pgfscope}%
\pgfsetbuttcap%
\pgfsetroundjoin%
\pgfsetlinewidth{1.204500pt}%
\definecolor{currentstroke}{rgb}{0.800000,0.447059,0.000000}%
\pgfsetstrokecolor{currentstroke}%
\pgfsetdash{}{0pt}%
\pgfpathmoveto{\pgfqpoint{1.508252in}{0.060000in}}%
\pgfpathlineto{\pgfqpoint{1.708252in}{0.060000in}}%
\pgfusepath{stroke}%
\end{pgfscope}%
\begin{pgfscope}%
\definecolor{textcolor}{rgb}{0.000000,0.000000,0.000000}%
\pgfsetstrokecolor{textcolor}%
\pgfsetfillcolor{textcolor}%
\pgftext[x=1.753252in,y=0.035000in,left,base]{\color{textcolor}{\rmfamily\fontsize{8.000000}{9.600000}\selectfont CR-FM-NES}}%
\end{pgfscope}%
\begin{pgfscope}%
\pgfsetbuttcap%
\pgfsetroundjoin%
\pgfsetlinewidth{1.204500pt}%
\definecolor{currentstroke}{rgb}{0.298039,0.352941,0.015686}%
\pgfsetstrokecolor{currentstroke}%
\pgfsetdash{}{0pt}%
\pgfpathmoveto{\pgfqpoint{2.515486in}{0.060000in}}%
\pgfpathlineto{\pgfqpoint{2.715486in}{0.060000in}}%
\pgfusepath{stroke}%
\end{pgfscope}%
\begin{pgfscope}%
\definecolor{textcolor}{rgb}{0.000000,0.000000,0.000000}%
\pgfsetstrokecolor{textcolor}%
\pgfsetfillcolor{textcolor}%
\pgftext[x=2.760486in,y=0.035000in,left,base]{\color{textcolor}{\rmfamily\fontsize{8.000000}{9.600000}\selectfont MR15-GA}}%
\end{pgfscope}%
\begin{pgfscope}%
\pgfsetbuttcap%
\pgfsetroundjoin%
\pgfsetlinewidth{1.204500pt}%
\definecolor{currentstroke}{rgb}{0.000000,0.000000,0.000000}%
\pgfsetstrokecolor{currentstroke}%
\pgfsetdash{}{0pt}%
\pgfpathmoveto{\pgfqpoint{3.422078in}{0.060000in}}%
\pgfpathlineto{\pgfqpoint{3.622078in}{0.060000in}}%
\pgfusepath{stroke}%
\end{pgfscope}%
\begin{pgfscope}%
\definecolor{textcolor}{rgb}{0.000000,0.000000,0.000000}%
\pgfsetstrokecolor{textcolor}%
\pgfsetfillcolor{textcolor}%
\pgftext[x=3.667078in,y=0.035000in,left,base]{\color{textcolor}{\rmfamily\fontsize{8.000000}{9.600000}\selectfont RL2CO}}%
\end{pgfscope}%
\begin{pgfscope}%
\pgfsetbuttcap%
\pgfsetroundjoin%
\pgfsetlinewidth{1.204500pt}%
\definecolor{currentstroke}{rgb}{0.694118,0.125490,0.513725}%
\pgfsetstrokecolor{currentstroke}%
\pgfsetdash{{4.440000pt}{1.920000pt}}{0pt}%
\pgfpathmoveto{\pgfqpoint{4.197211in}{0.060000in}}%
\pgfpathlineto{\pgfqpoint{4.397211in}{0.060000in}}%
\pgfusepath{stroke}%
\end{pgfscope}%
\begin{pgfscope}%
\definecolor{textcolor}{rgb}{0.000000,0.000000,0.000000}%
\pgfsetstrokecolor{textcolor}%
\pgfsetfillcolor{textcolor}%
\pgftext[x=4.442211in,y=0.035000in,left,base]{\color{textcolor}{\rmfamily\fontsize{8.000000}{9.600000}\selectfont Random schedule}}%
\end{pgfscope}%
\begin{pgfscope}%
\pgfsetbuttcap%
\pgfsetroundjoin%
\pgfsetlinewidth{1.204500pt}%
\definecolor{currentstroke}{rgb}{0.533333,0.533333,0.533333}%
\pgfsetstrokecolor{currentstroke}%
\pgfsetdash{{1.200000pt}{1.980000pt}}{0pt}%
\pgfpathmoveto{\pgfqpoint{5.414045in}{0.060000in}}%
\pgfpathlineto{\pgfqpoint{5.614045in}{0.060000in}}%
\pgfusepath{stroke}%
\end{pgfscope}%
\begin{pgfscope}%
\definecolor{textcolor}{rgb}{0.000000,0.000000,0.000000}%
\pgfsetstrokecolor{textcolor}%
\pgfsetfillcolor{textcolor}%
\pgftext[x=5.659045in,y=0.035000in,left,base]{\color{textcolor}{\rmfamily\fontsize{8.000000}{9.600000}\selectfont Random search}}%
\end{pgfscope}%
\begin{pgfscope}%
\pgfsetbuttcap%
\pgfsetroundjoin%
\pgfsetlinewidth{1.204500pt}%
\definecolor{currentstroke}{rgb}{0.000000,0.000000,0.000000}%
\pgfsetstrokecolor{currentstroke}%
\pgfsetdash{{1.200000pt}{1.980000pt}}{0pt}%
\pgfpathmoveto{\pgfqpoint{6.544977in}{0.060000in}}%
\pgfpathlineto{\pgfqpoint{6.744977in}{0.060000in}}%
\pgfusepath{stroke}%
\end{pgfscope}%
\begin{pgfscope}%
\definecolor{textcolor}{rgb}{0.000000,0.000000,0.000000}%
\pgfsetstrokecolor{textcolor}%
\pgfsetfillcolor{textcolor}%
\pgftext[x=6.789977in,y=0.035000in,left,base]{\color{textcolor}{\rmfamily\fontsize{8.000000}{9.600000}\selectfont VBS}}%
\end{pgfscope}%
\end{pgfpicture}%
\makeatother%
\endgroup%
\\[3pt]
  \gfitem{media/griewank_rosenbrock_f8f22d.pgf}{2-D loss landscape of the Griewank--Rosenbrock function.}{fig:griewank2d_landscape}
  \gfitem{media/rtd_task_griewank.pgf}{Empirical runtime distributions}{fig:rtd_task_griewank2d}
  \gfitem{media/raster_task_griewank.pgf}{Optimizer schedules per realization}{fig:raster_griewank2d}
  \caption{The composite Griewank--Rosenbrock function F8F2~\autocite{Hansen2009} in two dimensions.}
  \label{fig:griewank2d}
\end{gridfigure}

The same panel also shows three variants of a random schedule, which picks an optimizer uniformly at random at each decision point. They differ only in the duration of the commitment $\tau$. Initially, all three fall short of random search: early arbitrary choices rarely match the problem, and the schedule repeatedly pays a switching cost before benefiting from a good choice. At the end of the budget, two schedules surpass the strongest single optimizer in the targets reached, while the shortest commitment, $\tau = 10$, ends just below it. Shorter commitments create more decision points, and although the handshake transfers some information across switches, most of the loss landscape information is discarded each time. The random schedule is the learned policy's baseline: it matches the portfolio and switching cost, differing only by making each optimizer choice randomly.

\subsection{Evaluating the Learned Policy}
\label{subsec:evaluation}

\noindent We train the RL2CO model according to the parameters in Table~\ref{tab:training} on the training set from Section~\ref{subsec:taskset}. Figure~\ref{fig:rtd_bbob_headroom_test} shows the ERTDs on the held-out test set. The trained RL2CO policy reaches a larger fraction of targets than any member of the portfolio, and does so at all but the smallest budgets: only below roughly one evaluation per dimension does L-BFGS lead. Since the portfolio was constructed so that no optimizer dominates, the margin over all four static optimizers is attributable to the schedule. The gap to the random schedule attributes that margin to informed decisions rather than to switching itself. Appendix~\ref{app:test-set-breakdown} partitions the same test set by objective stochasticity and by problem dimensionality (Fig.~\ref{fig:rtd_groups}); the leading member differs between the subsets, yet the policy ends ahead within each of them. The advantage is not an artifact of averaging over subsets on which different optimizers lead.

By combining the exploratory tendencies of one optimizer with the exploitative strengths of another, the model can reach the targets faster than the VBS. Figure~\ref{fig:griewank2d} illustrates this behavior on the two-dimensional Griewank--Rosenbrock function F8F2~\autocite{Hansen2009}, whose loss surface (Fig.~\ref{fig:griewank2d_landscape}) contains a wide, shallow basin with numerous local optima. The runtime distributions are provided in Fig.~\ref{fig:rtd_task_griewank2d}; the schedules for individual realizations are shown in Fig.~\ref{fig:raster_griewank2d}, where color encodes which optimizer is used across the budget. The learned policy reaches the targets sooner than any single optimizer: it starts with an exploration stage with CR-FM-NES to locate the basin containing the global minimum and then switches to L-BFGS, which converges to the minimum using far fewer evaluations than continued global search would require.

We evaluate the trained policy on three further benchmark suites: the CEC 2005 special session on real-parameter optimization~\autocite{Suganthan2005} in Fig.~\ref{fig:rtd_cec2005}, the CEC 2017 competition on single-objective real-parameter optimization~\autocite{Wu2017a} in Fig.~\ref{fig:rtd_cec2017}, and the CEC 2013 special session on large-scale global optimization~\autocite{Li2013} in Fig.~\ref{fig:rtd_cec2013lsgo}. These suites vary in how much benefit an algorithm-selection policy can extract. For CEC 2005, the VBS--SBS gap is substantial, and the policy takes advantage of it: it finishes ahead of every portfolio constituent, although the separation becomes clear only late along the budget axis. In contrast, for CEC 2017 and CEC 2013 the gap is small and the policy matches the random schedule. When there is little exploitable selection signal, the policy reverts to its baseline behavior, which on both suites continues to be much better than random search and close to the single best solver. In this setting, losing the selection signal removes the edge of the policy over the portfolio, but not its practical value. This is exactly the kind of robustness that a schedule should have under a distribution shift.

\begin{gridfigure}[cols=3,sharey=false,gutter=0.06in,rowsep=1.5ex,asfigures=true,placement=t]
  \begingroup%
\makeatletter%
\begin{pgfpicture}%
\pgfpathrectangle{\pgfpointorigin}{\pgfqpoint{6.997095in}{0.140000in}}%
\pgfusepath{use as bounding box, clip}%
\begin{pgfscope}%
\pgfsetbuttcap%
\pgfsetroundjoin%
\pgfsetlinewidth{1.204500pt}%
\definecolor{currentstroke}{rgb}{0.039216,0.627451,0.823529}%
\pgfsetstrokecolor{currentstroke}%
\pgfsetdash{}{0pt}%
\pgfpathmoveto{\pgfqpoint{0.000000in}{0.060000in}}%
\pgfpathlineto{\pgfqpoint{0.200000in}{0.060000in}}%
\pgfusepath{stroke}%
\end{pgfscope}%
\begin{pgfscope}%
\definecolor{textcolor}{rgb}{0.000000,0.000000,0.000000}%
\pgfsetstrokecolor{textcolor}%
\pgfsetfillcolor{textcolor}%
\pgftext[x=0.245000in,y=0.035000in,left,base]{\color{textcolor}{\rmfamily\fontsize{8.000000}{9.600000}\selectfont L-BFGS}}%
\end{pgfscope}%
\begin{pgfscope}%
\pgfsetbuttcap%
\pgfsetroundjoin%
\pgfsetlinewidth{1.204500pt}%
\definecolor{currentstroke}{rgb}{0.368627,0.337255,0.870588}%
\pgfsetstrokecolor{currentstroke}%
\pgfsetdash{}{0pt}%
\pgfpathmoveto{\pgfqpoint{0.806557in}{0.060000in}}%
\pgfpathlineto{\pgfqpoint{1.006557in}{0.060000in}}%
\pgfusepath{stroke}%
\end{pgfscope}%
\begin{pgfscope}%
\definecolor{textcolor}{rgb}{0.000000,0.000000,0.000000}%
\pgfsetstrokecolor{textcolor}%
\pgfsetfillcolor{textcolor}%
\pgftext[x=1.051557in,y=0.035000in,left,base]{\color{textcolor}{\rmfamily\fontsize{8.000000}{9.600000}\selectfont Rprop}}%
\end{pgfscope}%
\begin{pgfscope}%
\pgfsetbuttcap%
\pgfsetroundjoin%
\pgfsetlinewidth{1.204500pt}%
\definecolor{currentstroke}{rgb}{0.800000,0.447059,0.000000}%
\pgfsetstrokecolor{currentstroke}%
\pgfsetdash{}{0pt}%
\pgfpathmoveto{\pgfqpoint{1.508252in}{0.060000in}}%
\pgfpathlineto{\pgfqpoint{1.708252in}{0.060000in}}%
\pgfusepath{stroke}%
\end{pgfscope}%
\begin{pgfscope}%
\definecolor{textcolor}{rgb}{0.000000,0.000000,0.000000}%
\pgfsetstrokecolor{textcolor}%
\pgfsetfillcolor{textcolor}%
\pgftext[x=1.753252in,y=0.035000in,left,base]{\color{textcolor}{\rmfamily\fontsize{8.000000}{9.600000}\selectfont CR-FM-NES}}%
\end{pgfscope}%
\begin{pgfscope}%
\pgfsetbuttcap%
\pgfsetroundjoin%
\pgfsetlinewidth{1.204500pt}%
\definecolor{currentstroke}{rgb}{0.298039,0.352941,0.015686}%
\pgfsetstrokecolor{currentstroke}%
\pgfsetdash{}{0pt}%
\pgfpathmoveto{\pgfqpoint{2.515486in}{0.060000in}}%
\pgfpathlineto{\pgfqpoint{2.715486in}{0.060000in}}%
\pgfusepath{stroke}%
\end{pgfscope}%
\begin{pgfscope}%
\definecolor{textcolor}{rgb}{0.000000,0.000000,0.000000}%
\pgfsetstrokecolor{textcolor}%
\pgfsetfillcolor{textcolor}%
\pgftext[x=2.760486in,y=0.035000in,left,base]{\color{textcolor}{\rmfamily\fontsize{8.000000}{9.600000}\selectfont MR15-GA}}%
\end{pgfscope}%
\begin{pgfscope}%
\pgfsetbuttcap%
\pgfsetroundjoin%
\pgfsetlinewidth{1.204500pt}%
\definecolor{currentstroke}{rgb}{0.000000,0.000000,0.000000}%
\pgfsetstrokecolor{currentstroke}%
\pgfsetdash{}{0pt}%
\pgfpathmoveto{\pgfqpoint{3.422078in}{0.060000in}}%
\pgfpathlineto{\pgfqpoint{3.622078in}{0.060000in}}%
\pgfusepath{stroke}%
\end{pgfscope}%
\begin{pgfscope}%
\definecolor{textcolor}{rgb}{0.000000,0.000000,0.000000}%
\pgfsetstrokecolor{textcolor}%
\pgfsetfillcolor{textcolor}%
\pgftext[x=3.667078in,y=0.035000in,left,base]{\color{textcolor}{\rmfamily\fontsize{8.000000}{9.600000}\selectfont RL2CO}}%
\end{pgfscope}%
\begin{pgfscope}%
\pgfsetbuttcap%
\pgfsetroundjoin%
\pgfsetlinewidth{1.204500pt}%
\definecolor{currentstroke}{rgb}{0.694118,0.125490,0.513725}%
\pgfsetstrokecolor{currentstroke}%
\pgfsetdash{{4.440000pt}{1.920000pt}}{0pt}%
\pgfpathmoveto{\pgfqpoint{4.197211in}{0.060000in}}%
\pgfpathlineto{\pgfqpoint{4.397211in}{0.060000in}}%
\pgfusepath{stroke}%
\end{pgfscope}%
\begin{pgfscope}%
\definecolor{textcolor}{rgb}{0.000000,0.000000,0.000000}%
\pgfsetstrokecolor{textcolor}%
\pgfsetfillcolor{textcolor}%
\pgftext[x=4.442211in,y=0.035000in,left,base]{\color{textcolor}{\rmfamily\fontsize{8.000000}{9.600000}\selectfont Random schedule}}%
\end{pgfscope}%
\begin{pgfscope}%
\pgfsetbuttcap%
\pgfsetroundjoin%
\pgfsetlinewidth{1.204500pt}%
\definecolor{currentstroke}{rgb}{0.533333,0.533333,0.533333}%
\pgfsetstrokecolor{currentstroke}%
\pgfsetdash{{1.200000pt}{1.980000pt}}{0pt}%
\pgfpathmoveto{\pgfqpoint{5.414045in}{0.060000in}}%
\pgfpathlineto{\pgfqpoint{5.614045in}{0.060000in}}%
\pgfusepath{stroke}%
\end{pgfscope}%
\begin{pgfscope}%
\definecolor{textcolor}{rgb}{0.000000,0.000000,0.000000}%
\pgfsetstrokecolor{textcolor}%
\pgfsetfillcolor{textcolor}%
\pgftext[x=5.659045in,y=0.035000in,left,base]{\color{textcolor}{\rmfamily\fontsize{8.000000}{9.600000}\selectfont Random search}}%
\end{pgfscope}%
\begin{pgfscope}%
\pgfsetbuttcap%
\pgfsetroundjoin%
\pgfsetlinewidth{1.204500pt}%
\definecolor{currentstroke}{rgb}{0.000000,0.000000,0.000000}%
\pgfsetstrokecolor{currentstroke}%
\pgfsetdash{{1.200000pt}{1.980000pt}}{0pt}%
\pgfpathmoveto{\pgfqpoint{6.544977in}{0.060000in}}%
\pgfpathlineto{\pgfqpoint{6.744977in}{0.060000in}}%
\pgfusepath{stroke}%
\end{pgfscope}%
\begin{pgfscope}%
\definecolor{textcolor}{rgb}{0.000000,0.000000,0.000000}%
\pgfsetstrokecolor{textcolor}%
\pgfsetfillcolor{textcolor}%
\pgftext[x=6.789977in,y=0.035000in,left,base]{\color{textcolor}{\rmfamily\fontsize{8.000000}{9.600000}\selectfont VBS}}%
\end{pgfscope}%
\end{pgfpicture}%
\makeatother%
\endgroup%
\\[3pt]
  \gfitem{media/ertd/rtd_cec2005.pgf}{Empirical runtime distributions on the CEC 2005 functions~\autocite{Suganthan2005}.}{fig:rtd_cec2005}
  \gfitem{media/ertd/rtd_cec2017.pgf}{Empirical runtime distributions on the CEC 2017 functions~\autocite{Wu2017a}.}{fig:rtd_cec2017}
  \gfitem{media/ertd/rtd_cec2013lsgo.pgf}{Empirical runtime distributions on the CEC 2013 large-scale global optimization functions~\autocite{Li2013}.}{fig:rtd_cec2013lsgo}
\end{gridfigure}

\subsection{Transfer to Unseen Problem Families}
\label{subsec:transfer}

\noindent The evaluation in Section~\ref{subsec:evaluation} uses analytic benchmark functions. We assess the learned policy on three sets of tasks from unseen problem families. The first consists of randomly sampled convex quadratic functions shown in Fig.~\ref{fig:rtd_quadratic}. The second involves training a small neural network on synthetic Gaussian data, as depicted in Fig.~\ref{fig:rtd_gaussian_classification}. The third considers a meta-optimization task with three parameters matching Adam's hyperparameters, where the objective is the loss obtained by a small classifier after a fixed inner training run in Fig.~\ref{fig:rtd_gaussian_meta}. Figure~\ref{fig:raster_tasks} shows the optimizer schedules and Appendix~\ref{app:families} gives the exact definitions of all three tasks.

\begin{gridfigure}[cols=3,sharey=false,gutter=0.06in,placement=!t]
  \gfitem{media/ertd/rtd_quadratic.pgf}{Randomly sampled convex quadratics.}{fig:rtd_quadratic}
  \gfitem{media/ertd/rtd_gaussian_classification.pgf}{Gaussian classification tasks.}{fig:rtd_gaussian_classification}
  \gfitem{media/ertd/rtd_gaussian_meta.pgf}{Meta-optimization tasks.}{fig:rtd_gaussian_meta}
  \caption{Empirical runtime distributions on the three unseen problem families. Each panel carries the four portfolio optimizers, the learned policy, a random schedule over the same optimizers, random search, and the VBS.}
  \label{fig:rtd_families}
\end{gridfigure}

\begin{gridfigure}[cols=3,sharey=false,gutter=0.06in,placement=!t]
  \gfitem{media/raster_quadratics.pgf}{Randomly sampled convex quadratics.}{fig:raster_quadratics}
  \gfitem{media/raster_gaussian_classification.pgf}{Gaussian classification tasks.}{fig:raster_gaussian_classification}
  \gfitem{media/raster_gaussian_meta.pgf}{Meta-optimization tasks.}{fig:raster_gaussian_meta}
  \caption{Optimizer schedules the learned policy ran on the three unseen problem families, one row per task and realization, over the same budget axis as their runtime distributions in Fig.~\ref{fig:rtd_families}. Color marks which optimizer the policy used at that point of the budget.}
  \label{fig:raster_tasks}
\end{gridfigure}

The three task families differ in whether one optimizer dominates and whether the policy recognizes that dominance. For the quadratics, the objective is convex, L-BFGS clearly wins, and the VBS reduces to that choice. The policy detects this and acts as a static selector. For Gaussian classification, Rprop is fastest early, but CR-FM-NES wins at higher budgets, where low dimensionality favors longer-horizon evolution. The policy settles with MR15-GA initially, but recovers after switching to Rprop and still reaches the targets on an unseen family. For meta-optimization, all optimizers hit the targets, but CR-FM-NES is about an order of magnitude faster than the gradient-based methods, matching the irregular inner landscape, and the policy is faster than any single method. Overall, the policy exploits the selection structure when present and otherwise falls back on the behavior of the random schedule, surpassing the performance of the single optimizers given enough budget.

\section{Conclusion}
\label{sec:conclusion}

\noindent We introduced RL2CO, which treats the evolution of an optimization run as a sequential decision-making problem. A recurrent policy selects the next optimizer to run and how many iterations it may take. A switching mechanism transfers the current candidate solution and the search-scale information to the incoming optimizer. A context proxy conditions a gating network that routes among multiple experts. We designed the portfolio and training tasks jointly so that no single algorithm is sufficiently strong to act as a default choice. On the unseen benchmark function set, the learned policy achieves a higher target-hit rate, at all but the smallest budgets, than any individual optimizer it coordinates. On a broad collection of test problems, what the policy gains depends on how much an informed choice can still deliver there. Where that margin is wide, the policy finishes ahead of every portfolio member, although the separation becomes visible only late along the budget axis. Where it is narrow, the policy reverts to the behavior of its random baseline, which ends well above a random search and close to the strongest single optimizer.

A switch currently transfers the incumbent solution and a scalar search scale, while the rest of each member's internal state is kept or rebuilt on its own side of the family boundary. Learning a translation between these internal representations could carry the search structure accumulated by one member across that boundary. A commitment's duration is also measured in iterations, yet portfolio members differ by orders of magnitude in the number of evaluations one iteration entails; expressing durations in evaluations, or in wall-clock time, would make that trade-off explicit to the policy. The evaluation so far stays on analytic benchmarks and small learning tasks, and exercising the policy on simulation-based design problems, where every evaluation carries real cost, is the natural next step. A limitation of this model is that the policy generalizes across task distributions but not across portfolios. Because the optimizer choice is a discrete action, portfolio members are fixed during training. Changing or adding optimizers alters the action space and requires retraining the policy from scratch.

\section*{Acknowledgments}

\noindent This work is supported by the TU Delft AI Labs \& Talent Programme. The first author acknowledges the use of large language models to edit the text of this manuscript after a complete draft was written.

\appendix
\makeatletter
\let\@seccntformat\rl@appendixseccntformat
\makeatother

\section{Environment Details}
\label{app:environment}

\noindent The observation $o_t$ is constructed by concatenating five components from state $s_t$: (1) the mean loss per iteration, (2) the best loss observed so far, (3) a one-hot encoding of the optimizer used in that iteration, (4) the best loss each optimizer has reached separately, and (5) the fraction of the total function evaluation budget used in that iteration. The loss components are normalized and log-transformed relative to the initial loss and the task-specific global minimum.

Given $s_t$ and $a_t$, the transition passes control to $g_{k_t}$ via an optimizer handshake, described in more detail in Appendix~\ref{app:handshake}. That optimizer then runs for $\tau_t$ iterations, and the resulting losses are appended to the history. During training, an episode ends when the iteration budget $B$ is exhausted, when the global minimum is reached, or when the decision horizon $H$ is spent. At deployment, the policy can be consulted as long as budget remains.

The reward depends on how much the chosen action improves performance and how much of the remaining budget can still be used to exploit that improvement. Let $\nu(s)$ be the number of function evaluations used in state $s$, $\nu_{\max}$ the total evaluation budget, and $h(s)$ the best objective value observed so far, normalized and log-transformed as in $o_t$. Because anytime performance is measured on a logarithmic evaluation scale, define $\Lambda(s)$ as the integral of $h$ along this axis up to $\nu(s)$. The potential

\begin{equation}
    \Phi(s) = \Lambda(s) + h(s)\left[\log(1 + \nu_{\max}) - \log(1 + \nu(s))\right]
    \label{eq:potential}
\end{equation}

is the total the run would be charged if it stalled at $s$ and never improved again, and the reward is the normalized decrease of that potential over a decision:

\begin{equation}
    R_c(s, a) = \frac{\Phi(s) - \Phi(s')}{\log(1 + \nu_{\max})}
    \label{eq:reward}
\end{equation}

with $s'$ denoting the state reached after taking the decision. Normalization ensures that the undiscounted total reward over an episode lies in $[0, 1]$ for each task, so every task contributes equally during training.

\subsection{The Handshake}
\label{app:handshake}

\noindent Switching to an optimizer different from the one used in the preceding iteration initiates a \textit{handshake}: data from the outgoing optimization trajectory are converted to initialize the internal state parameters of the new optimizer. Most optimizers implicitly follow a convergence schedule: they begin with a wide exploratory search and gradually focus on a promising region of the loss landscape. If a policy switches optimizers mid-run and initializes the new optimizer's state from scratch, the search scale resets to this global level, discarding the fine-grained resolution already achieved. To prevent this, we transfer a scalar $\sigma$ across optimizer switches, which acts as a characteristic step size: with $d$ the dimension of the problem, $d \cdot \sigma^2$ is the expected squared distance per iteration. Each optimizer defines a contract to derive $\sigma$ from its internal state; the contracts in our optimizer portfolio appear in Table~\ref{tab:sigma_construction}.

\begin{table}[!t]
\caption{Construction of the Transferred Scale $\sigma$.}
\label{tab:sigma_construction}
\centering
\footnotesize
\begin{tabularx}{\columnwidth}{l >{\hsize=0.75\hsize\raggedright\arraybackslash}X >{\hsize=1.25\hsize\raggedright\arraybackslash}X}
\toprule
\textbf{Optimizer} & \textbf{State read} & \textbf{Scale $\sigma$} \\
\midrule
L-BFGS    & last accepted step $\Delta \boldsymbol{x}$                                & $\lVert \Delta \boldsymbol{x} \rVert / \sqrt{d}$ \\
          & before one exists, the gradient $\boldsymbol{g}$ and the accepted step length $\alpha$ & $\alpha \min(\lVert \boldsymbol{g} \rVert, 1) / \sqrt{d}$ \\
\addlinespace
Rprop     & per-coordinate step sizes $\boldsymbol{\Delta}$                           & $\lVert \boldsymbol{\Delta} \rVert / \sqrt{d}$ \\
\addlinespace
CR-FM-NES & sampling width $s$ and shape fields $\boldsymbol{D}$, $\boldsymbol{v}$    & $s \sqrt{d^{-1} \sum_{i} D_{i}^{2} (1 + v_{i}^{2})}$ \\
\addlinespace
MR15-GA   & mutation width $s$                                                        & $s$ \\
\bottomrule
\end{tabularx}
\end{table}

The state of an incoming optimizer is reconstructed from four quantities: the best candidate found so far $\boldsymbol{x}^{\star}$, its objective function value $f(\boldsymbol{x}^{\star})$, the scale parameter $\sigma$ and the dimensionality $d$. The transfer is carried out in three stages. First, the internal state of the incoming optimizer is reset or resumed. Only for L-BFGS is the internal state reset, since the pairs of step and gradient difference it stores approximate the curvature of the loss along the iterates that produced them, and that approximation says nothing about the curvature at $\boldsymbol{x}^{\star}$. Next, the state is modified by the three properties. Lastly, a new starting point or population is initialized. CR-FM-NES and MR15-GA generate a new population from their modified state. For Rprop and L-BFGS, the best candidate is suggested as a starting point. Table~\ref{tab:handshake} states the procedure for all three steps, for each of the four portfolio optimizers.

\begin{table}[!t]
\caption{The Three Steps of the Handshake.}
\label{tab:handshake}
\centering
\footnotesize
\setlength{\tabcolsep}{4pt}
\begin{tabularx}{\columnwidth}{l c >{\hsize=1.25\hsize\raggedright\arraybackslash}X >{\hsize=0.75\hsize\raggedright\arraybackslash}X}
\toprule
\textbf{Optimizer} & \textbf{Reset} & \textbf{Internal parameters written} & \textbf{Next candidates} \\
\midrule
L-BFGS & Yes & none & set to $\boldsymbol{x}^{\star}$ \\
\addlinespace
Rprop & No & $\boldsymbol{\Delta} \leftarrow \sqrt{d}\, \sigma \boldsymbol{\Delta} / \lVert \boldsymbol{\Delta} \rVert$ \newline sign memory $\leftarrow \boldsymbol{0}$ & set to $\boldsymbol{x}^{\star}$ \\
\addlinespace
CR-FM-NES & No & $\boldsymbol{m} \leftarrow \boldsymbol{x}^{\star}$ \newline $s \leftarrow \sigma / \sqrt{d^{-1} \sum_{i} D_{i}^{2} (1 + v_{i}^{2})}$ \newline evolution paths $\boldsymbol{p}_{s}, \boldsymbol{p}_{c} \leftarrow \boldsymbol{0}$, the shape fields $\boldsymbol{D}$, $\boldsymbol{v}$ kept & generated from the state \\
\addlinespace
MR15-GA & No & $s \leftarrow \sigma$ \newline every elite row $\boldsymbol{e}_{i} \leftarrow \boldsymbol{x}^{\star}$, and the fitness stored with it $\leftarrow f(\boldsymbol{x}^{\star})$ & generated from the state \\
\bottomrule
\end{tabularx}
\end{table}

\section{Training Details}
\label{app:training}

\noindent DAAC training proceeds in rounds. In each round, we sample a subset of training problems and run rollouts under the current policy. We then update the policy network with minibatches, while updating the value network every few rounds. Algorithm~\ref{alg:training} summarizes the procedure.

\begin{algorithm}[!t]
\caption{Training the Schedule Policy\label{alg:training}}
\begin{algorithmic}[1]
\Require training contexts $\mathcal{C}_{\text{train}}$, rounds $N$, $m$ contexts
    and $L$ realizations per round
\Require $E_\pi$, $E_V$, $N_V$, $\epsilon$, $\alpha_s$, $\alpha_a$,
    $\alpha_{\text{lb}}$, $\gamma$, $\lambda$, $\eta$
\State initialize policy $\boldsymbol{\theta}$ and value $\boldsymbol{\phi}$
\For{$n = 1, \dots, N$}
    \State $\mathcal{C}_n \gets m$ contexts sampled from $\mathcal{C}_{\text{train}}$
    \For{$c \in \mathcal{C}_n$ \textbf{in parallel}}
        \State $\{\zeta_{c,\ell}\}_{\ell=1}^{L} \gets L$ episodes of $\pi_{\boldsymbol{\theta}}$ on $c$
            \Comment{$\zeta = (o_t, a_t, r_t, \mathrm{done}_t, v_t, \Delta_t)_{t \le H}$}
    \EndFor
    \State $\mathcal{D} \gets \bigcup_{c,\ell} \zeta_{c,\ell}$ \Comment{pool across contexts}
    \For{$t = H, \dots, 0$ over every $\zeta \in \mathcal{D}$}
        \State $\delta_t \gets r_t + \gamma^{\Delta_t} v_{t+1} (1 - \mathrm{done}_t) - v_t$
            \Comment{iterations consumed}
        \State $\hat{A}_t \gets \delta_t + \gamma^{\Delta_t} \lambda (1 - \mathrm{done}_t) \hat{A}_{t+1}$
        \State $\hat{G}_t \gets \hat{A}_t + v_t$
    \EndFor
    \State $\hat{A} \gets (\hat{A} - \mu_{\mathcal{D}}) / \sigma_{\mathcal{D}}$
        \Comment{over the pooled batch}
    \State $\boldsymbol{\theta}_{\text{old}} \gets \boldsymbol{\theta}$
    \For{$E_\pi$ epochs, minibatch $\mathcal{M} \subseteq \mathcal{D}$}
        \State $\rho_t \gets \pi_{\boldsymbol{\theta}}(a_t \mid o_t) \,/\,
            \pi_{\boldsymbol{\theta}_{\text{old}}}(a_t \mid o_t)$
        \State $J_\pi \gets \mathbb{E}_{\mathcal{M}}\big[\min\big(\rho_t \hat{A}_t,$
        \Statex \hskip\algorithmicindent\hskip\algorithmicindent
            $\operatorname{clip}(\rho_t, 1 \!-\! \epsilon, 1 \!+\! \epsilon)\,
            \hat{A}_t\big)\big]$
        \State $L_A \gets \mathbb{E}_{\mathcal{M}}\big[(A_{\boldsymbol{\theta}}(o_t, a_t)
            - \hat{A}_t)^2\big]$
        \State $J \gets J_\pi + \alpha_s \mathcal{H}[\pi_{\boldsymbol{\theta}}] - \alpha_a L_A$
            \Comment{entropy of both heads}
        \If{mixture arm}
            \State $J \gets J - \alpha_{\text{lb}} K_{\text{e}} \sum_{j} \bar{p}_j^2$
                \Comment{$\bar{p}_j$: mean gate weight}
        \EndIf
        \State $\boldsymbol{\theta} \gets \boldsymbol{\theta}
            + \eta \nabla_{\boldsymbol{\theta}} J$
    \EndFor
    \If{$n \bmod N_V = 0$}
        \For{$E_V$ epochs, minibatch $\mathcal{M} \subseteq \mathcal{D}$}
            \State $\boldsymbol{\phi} \gets \boldsymbol{\phi} - \eta \nabla_{\boldsymbol{\phi}}
                \mathbb{E}_{\mathcal{M}}\big[(V_{\boldsymbol{\phi}}(o_t) - \hat{G}_t)^2\big]$
        \EndFor
    \EndIf
\EndFor
\State \Return $\boldsymbol{\theta}$
\end{algorithmic}
\end{algorithm}

The inputs are the training context set $\mathcal{C}_{\text{train}}$, the number of rounds $N$, the number of contexts $m$ sampled per round, and the number of independent realizations $L$, so each rollout produces $mL$ episodes. The hyperparameters are the epochs $E_\pi$ and $E_V$ per policy and value update, value-update period $N_V$ (in rounds), clipping half-width $\epsilon$, and coefficients $\alpha_s$, $\alpha_a$, and $\alpha_{\text{lb}}$ weighting the entropy bonus, auxiliary advantage loss, and load-balancing penalty. $\gamma$ and $\lambda$ are the discount factor and GAE trace parameter. The learnable parameters are policy-network weights $\boldsymbol{\theta}$ for $\pi_{\boldsymbol{\theta}}$ (including the auxiliary advantage head $A_{\boldsymbol{\theta}}$) and value-network weights $\boldsymbol{\phi}$ for $V_{\boldsymbol{\phi}}$; $\eta$ is the Adam learning rate.

In round $n$, we sample a subset $\mathcal{C}_n \subseteq \mathcal{C}_{\text{train}}$ of $m$ contexts and, for each $c \in \mathcal{C}_n$, run $L$ episodes $\zeta_{c,\ell}$ with the current policy. Each trajectory $\zeta$ records at every decision step $t$: observation $o_t$, action $a_t$, reward $r_t$, termination flag $\mathrm{done}_t$, value prediction $v_t = V_{\boldsymbol{\phi}}(o_t)$, and the number of optimizer iterations $\Delta_t$ used by that decision. All episodes are then divided into minibatches $\mathcal{M} \subseteq \mathcal{D}$.

The backward pass over each trajectory computes the temporal-difference residual $\delta_t$, the generalized advantage estimate $\hat{A}_t$, and the return target $\hat{G}_t$. The advantage estimates are then normalized using the batch mean $\mu_{\mathcal{D}}$ and standard deviation $\sigma_{\mathcal{D}}$.

The clipped surrogate objective $J_\pi$ is the PPO loss: the empirical minibatch mean $\mathbb{E}_{\mathcal{M}}$ of the minimum between the unclipped and clipped ratios, where $\operatorname{clip}$ constrains $\rho_t$ to $[1 - \epsilon, 1 + \epsilon]$. The auxiliary loss $L_A$ trains the advantage head $A_{\boldsymbol{\theta}}(o_t, a_t)$, which reads the policy latent together with learned embeddings of the taken action, to match the target advantages $\hat{A}_t$, and $\mathcal{H}[\pi_{\boldsymbol{\theta}}]$ is the policy entropy over both optimizer and duration heads. These three terms together define the overall ascent objective $J$.

In the mixture arm, $J$ includes a load-balancing regularizer over the $K_{\text{e}}$ experts, where $\bar{p}_j$ is the average gate weight for expert $j$ over the minibatch. The term $K_{\text{e}} \sum_j \bar{p}_j^2$ attains its minimum value of one when probability mass is distributed uniformly across experts and increases as it concentrates on a single expert. The value function network is trained on the same batch every $N_V$ rounds by minimizing the squared error between $V_{\boldsymbol{\phi}}(o_t)$ and $\hat{G}_t$, which decouples the critic from the policy representation.

Because the process is partially observable, the policy depends on the observation history, not just the latest observation. We approximate this history with the hidden state of a recurrent network, carried across decisions and reset between episodes. Recurrence is the standard response to partial observability in model-free reinforcement learning~\autocite{Hausknecht2009}, and remains a strong baseline against explicit belief-state methods~\autocite{Ni2022}. In meta-optimization it is already the default way to handle temporal dependence; \textcite{Gomes2021} use it in a partially observable formulation of learned population-based optimizers. However, recurrent meta-policies train poorly over horizons of hundreds of generations~\autocite{Ma2025}. Here this problem is much reduced: one decision commits many iterations, so an episode spans at most $H$ decisions instead of thousands of iterations.

Table~\ref{tab:training} gives the dimensions of both networks and the values behind every symbol the algorithm names.

\begin{table}[!t]
\caption{Network Dimensions and Training Hyperparameters. Layer sizes read input $\to$ hidden $\to$
output.\label{tab:training}}
\centering
\footnotesize
\begin{tabularx}{\columnwidth}{L c c}
\toprule
\textbf{Quantity} & \textbf{Symbol} & \textbf{Value} \\
\midrule
\multicolumn{3}{l}{\textit{Trunk, one untied copy per network}} \\
Experts                  & $K_{\text{e}}$                               & $4$ \\
History encoder (long short-term memory)   &                                              & $6 \to 16 \to 8$ \\
Context descriptors      & $\tilde{c}$                                  & $3$ \\
Descriptor embedding     &                                              & $4$ \\
Context projection       &                                              & $12 \to 4$ \\
Progress embedding       &                                              & $2 \to 4$ \\
Latent projection        & $z_t$                                        & $16 \to 8$ \\
\addlinespace
\multicolumn{3}{l}{\textit{Heads}} \\
Optimizer                & $\pi_{\boldsymbol{\theta}}(k_t \mid o_t)$    & $8 \to 64 \to 64 \to 4$ \\
Duration                 & $\pi_{\boldsymbol{\theta}}(\tau_t \mid o_t)$ & $8 \to 64 \to 64 \to 4$ \\
Advantage                & $A_{\boldsymbol{\theta}}(o_t, a_t)$          & $24 \to 64 \to 64 \to 1$ \\
Critic                   & $V_{\boldsymbol{\phi}}(o_t)$                 & $8 \to 256 \to 256 \to 1$ \\
Optimizer embedding      & $\mathrm{embed}(k_t)$                        & $4 \times 8$ \\
Duration embedding       & $\mathrm{embed}(\tau_t)$                     & $4 \times 8$ \\
Activation               &                                              & $\tanh$ \\
\addlinespace
\multicolumn{3}{l}{\textit{Episode}} \\
Decision horizon         & $H$                                          & $100$ \\
Durations                & $\mathcal{T}$                                & $\{10, 100, 1000\}$ \\
Portfolio size           & $K$                                          & $4$ \\
\addlinespace
\multicolumn{3}{l}{\textit{Training}} \\
Rounds                   & $N$                                          & $100$ \\
Contexts per round       & $m$                                          & $25$ \\
Realizations per context & $L$                                          & $25$ \\
Policy epochs            & $E_\pi$                                      & $10$ \\
Value epochs             & $E_V$                                        & $10$ \\
Value-update period      & $N_V$                                        & $5$ \\
Minibatches per epoch    &                                              & $4$ \\
Clipping half-width      & $\epsilon$                                   & $0.25$ \\
Entropy coefficient      & $\alpha_s$                                   & $0.01$ \\
Advantage coefficient    & $\alpha_a$                                   & $0.25$ \\
Balancing coefficient    & $\alpha_{\text{lb}}$                         & $0.01$ \\
Discount per iteration   & $\gamma$                                     & $0.9999$ \\
GAE trace                & $\lambda$                                    & $0.9$ \\
Optimizer                &                                              & Adam \\
Learning rate            & $\eta$                                       & $3 \times 10^{-4}$ \\
Learning-rate decay      &                                              & linear to $0$ \\
Gradient-norm clip       &                                              & $0.5$ \\
\bottomrule
\end{tabularx}
\end{table}

Figures~\ref{fig:training_return} and~\ref{fig:training_critic} report the cumulative return during training and the critic loss, respectively.

\begin{gridfigure}[cols=2,sharey=false,gutter=1pc,asfigures=true]
  \gfitem{media/health_return.pgf}{Mean cumulative return against policy updates. The training curve is the mean over the tasks sampled in each round; the validation curve is the mean over $100$ held-out tasks, evaluated every fifth round and starting from the untrained policy.}{fig:training_return}
  \gfitem{media/health_critic.pgf}{Critic loss of the training run against the policy updates, recorded only every $N_V$ rounds.}{fig:training_critic}
\end{gridfigure}

\section{Benchmark Construction}
\label{app:dataset}

\noindent This appendix describes how we produced the profiling pool from which the two task sets in Section~\ref{subsec:taskset} were drawn.

\subsection{Benchmark Functions}
\label{app:benchmark-functions}

\noindent We implement the noiseless BBOB suite~\autocite{Hansen2009} and its noisy counterpart~\autocite{Hansen2009a} following their original specifications, except that instance-defining translations and rotations are sampled from a seed instead of read from the original data files. The implementation, written in JAX for direct use with gradient-based optimizers via automatic differentiation, is publicly available\footnote{\url{https://github.com/bessagroup/bbob-jax}~\autocite{vanderSchelling2025}}.

The high-dimensional function family embeds a $d_{\text{int}}$-dimensional BBOB function $f$ in a $d$-dimensional space with $d_{\text{int}} \le d$:

\begin{equation}
    F(\boldsymbol{x}) = f\!\left(\boldsymbol{z}^{*} + w\,\boldsymbol{B}\,(\boldsymbol{x} - \boldsymbol{\xi})\right),
    \label{eq:embedded}
\end{equation}

for $\boldsymbol{x} \in [0,1]^{d}$. The matrix $\boldsymbol{B} \in \mathbb{R}^{d_{\text{int}} \times d}$ has orthonormal rows from the reduced QR factorization of a Gaussian matrix. The optimum of $f$ is $\boldsymbol{z}^{*}$, and $w$ is the width of the box constraints of $f$, ensuring that uniform sampling in the ambient box covers the intended input range. The anchor $\boldsymbol{\xi}$ is a seeded uniform draw from the box. The gradient of $F$ lies in the $d_{\text{int}}$-dimensional row space of $\boldsymbol{B}$, matching the subspace concentration observed for gradient descent on deep networks~\autocite{Gur-Ari2018}. The Hessian spectrum includes the $d_{\text{int}}$ eigenvalues of $f$, each scaled by $w^2$, together with $d-d_{\text{int}}$ zeros, yielding the bulk-plus-outliers shape reported for such models~\autocite{Sagun2018,Papyan2019}. Because $F$ depends on $\boldsymbol{x}$ only through $\boldsymbol{B}(\boldsymbol{x} - \boldsymbol{\xi})$, it is constant along the null space of $\boldsymbol{B}$, and the minimizer is a $(d-d_{\text{int}})$-dimensional flat manifold rather than a point. Table~\ref{tab:pool} lists the profiling pool of the three benchmark sets.

\begin{table}[!t]
\caption{The Benchmark Function Pool, With $D = \{2, 3, 5, 10, 20, 40\}$.}
\label{tab:pool}
\centering
\setlength{\tabcolsep}{4pt}
\begin{tabularx}{\columnwidth}{l c L c c}
\toprule
\textbf{Family} & \textbf{Functions} & \textbf{Dimensions} & \textbf{Seeds} & \textbf{Instances} \\
\midrule
\texttt{bbob}          & 24 & $d \in D$ & 4 & 576  \\
\texttt{bbob-embedded} & 24 & $d_{\text{int}} \in D$ \newline $d \in \{64, 256, 1024\}$ & 4 & 1728 \\
\texttt{bbob-noisy}    & 30 & $d \in D$ & 4 & 720  \\
\midrule
\textbf{Total}         &    &            &   & 3024 \\
\bottomrule
\end{tabularx}
\end{table}

Each benchmark function has a region of interest: the part of the domain containing the global minimum and the non-trivial structural features. We rescale all functions so this region matches the unit box. For each instance, we run optimization for up to $10^4 \times d$ iterations, capped at 25,000, over 25 independent runs with initial points $\boldsymbol{x}^0 \sim \mathcal{N}(\boldsymbol{0}, \boldsymbol{I})$. Empirical runtime distributions follow \textcite{Hansen2022} with simulated restarts, except that we replace the original targets with 51 logarithmically spaced values between the third quartile of a random-search baseline and $10^{-8}$ above the global minimum. Using a random-search-based easiest target rather than a fixed absolute error keeps the target scale informative for all functions.

\subsection{Candidate Optimizers}
\label{app:dataset-optimizers}

\noindent The candidate set includes 52 optimizers plus a random-search baseline: 28 gradient-based methods from the \texttt{optax} library~\autocite{Deepmind2020} and 24 derivative-free methods from the \texttt{evosax} library~\autocite{Lange2022}. Table~\ref{tab:hyperparameters} gives the configuration of the four optimizers named in Section~\ref{subsec:taskset}. For all remaining optimizers, we used the default hyperparameter settings provided by their respective libraries.

\begin{table}[!t]
\caption{Hyperparameters of the Algorithm Portfolio.}
\label{tab:hyperparameters}
\centering
\footnotesize
\begin{tabularx}{\columnwidth}{l L c}
\toprule
\textbf{Optimizer} & \textbf{Parameter} & \textbf{Value} \\
\midrule
L-BFGS    & memory size                 & $10$ \\
          & line-search trials          & $\le 20$ \\
          & Wolfe tolerances            & $10^{-4}$, $0.9$ \\
\addlinespace
Rprop     & learning rate               & $10^{-3}$ \\
          & step multipliers            & $0.5$, $1.2$ \\
          & step-size bounds            & $10^{-6}$, $50$ \\
\addlinespace
CR-FM-NES & population size             & $2\lceil \lfloor 4 + 3 \ln d \rfloor / 2 \rceil$ \\
          & initial standard deviation  & $1$ \\
          & mean, moving-$s$ rates      & $1$ \\
\addlinespace
MR15-GA   & population size             & $\lfloor 4 + 3 \ln d \rfloor$ \\
          & initial standard deviation  & $1$ \\
          & target success rate         & $0.2$ \\
          & $s$ multipliers             & $0.5$, $2$ \\
          & elite fraction              & $1/2$ \\
\bottomrule
\end{tabularx}
\end{table}

\subsection{Joint Problem and Portfolio Selection}
\label{app:selection-procedure}

\noindent The procedure jointly selects benchmark problems and a portfolio of optimizers so that algorithm selection is meaningful. A task set is informative only if no single optimizer suffices: portfolio members must be complementary, the best choice must vary by problem, and none should be strong enough to serve as a default. All three depend on the selected problems and optimizers.

Let $\mathcal{P}$ be the benchmark problems, $\mathcal{O}$ the candidate optimizers, and $u_{p,o}$ the performance of $o \in \mathcal{O}$ on $p \in \mathcal{P}$. Let $\boldsymbol{\chi} \in \{0,1\}^{|\mathcal{P}|}$ encode the selection of $P$ problems. For a portfolio $S \subseteq \mathcal{O}$ of $K$ optimizers, define $v_p = \max_{o \in S} u_{p,o}$ (virtual best solver on $p$) and $m_p = \frac{1}{K}\sum_{o \in S} u_{p,o}$ (portfolio average on $p$). The objective is the sum of $v_p - m_p$ over the selected problems, measuring the benefit of informed algorithm selection. We constrain the portfolio so that no single optimizer dominates: the best optimizer's mean performance $\max_{o \in S} \frac{1}{P}\sum_{p} u_{p,o}\chi_p$ may exceed the portfolio mean by at most $\varepsilon$. This constraint is what lets the portfolio average $m_p$ stand in for the single best solver of Section~\ref{subsec:taskset}, whose identity depends on the selection itself and would make the objective a maximum over optimizers rather than a per-problem constant. A further constraint ties $S$ to all candidates: on the selected problems, the virtual best over $S$ must outperform each $o \in \mathcal{O}$ by at least $\mu$. With $S$ fixed, all quantities are averages over the chosen instances and are linear in $\boldsymbol{\chi}$, making the optimization problem tractable.

\begin{equation}
\begin{aligned}
\max_{S,\,\boldsymbol{\chi}} \;\; & \sum_{p \in \mathcal{P}} (v_p - m_p)\,\chi_p \\
\text{s.t.} \;\; & \sum_{p \in \mathcal{P}} \chi_p = P, \\
& \sum_{p \in \mathcal{P}} (u_{p,o} - m_p)\,\chi_p \le P\varepsilon \quad \forall\, o \in S, \\
& \sum_{p \in \mathcal{P}} (v_p - u_{p,o})\,\chi_p \ge P\mu \quad \forall\, o \in \mathcal{O}.
\end{aligned}
\label{eq:selection}
\end{equation}

We instantiate the program with $P = 100$ instances and a portfolio of $K = 4$ optimizers, chosen from the 52 candidates of Appendix~\ref{app:dataset-optimizers}. We set $\varepsilon = \mu = 0.02$. Since~\eqref{eq:selection} is a binary linear program in $\boldsymbol{\chi}$, we solve the combined problem by enumerating all $\binom{52}{K}$ portfolios. For each portfolio, we first solve its linear relaxation to obtain an upper bound on the integer optimum. We rank the portfolios by this bound and then solve them exactly in descending order, stopping once the bound of the next portfolio is below the best integer solution found so far, discarding all remaining portfolios. The test set is formed by repeating this procedure on the remaining problems.

\subsection{ERTD Breakdown}
\label{app:test-set-breakdown}

\noindent Figure~\ref{fig:rtd_bbob_headroom_train} gives the training set's ERTD for reference. Figures~\ref{fig:rtd_dim_2}--\ref{fig:rtd_dim_1024} resolve the comparison of Section~\ref{subsec:evaluation} by problem dimensionality, one ERTD per group. Figures~\ref{fig:rtd_noise_deterministic} and~\ref{fig:rtd_noise_stochastic} split the test set by whether the objective is stochastic or not.

\begin{gridfigure}[cols=3,sharey=false,gutter=0.06in,placement=!tp]
  \begingroup%
\makeatletter%
\begin{pgfpicture}%
\pgfpathrectangle{\pgfpointorigin}{\pgfqpoint{6.997095in}{0.140000in}}%
\pgfusepath{use as bounding box, clip}%
\begin{pgfscope}%
\pgfsetbuttcap%
\pgfsetroundjoin%
\pgfsetlinewidth{1.204500pt}%
\definecolor{currentstroke}{rgb}{0.039216,0.627451,0.823529}%
\pgfsetstrokecolor{currentstroke}%
\pgfsetdash{}{0pt}%
\pgfpathmoveto{\pgfqpoint{0.000000in}{0.060000in}}%
\pgfpathlineto{\pgfqpoint{0.200000in}{0.060000in}}%
\pgfusepath{stroke}%
\end{pgfscope}%
\begin{pgfscope}%
\definecolor{textcolor}{rgb}{0.000000,0.000000,0.000000}%
\pgfsetstrokecolor{textcolor}%
\pgfsetfillcolor{textcolor}%
\pgftext[x=0.245000in,y=0.035000in,left,base]{\color{textcolor}{\rmfamily\fontsize{8.000000}{9.600000}\selectfont L-BFGS}}%
\end{pgfscope}%
\begin{pgfscope}%
\pgfsetbuttcap%
\pgfsetroundjoin%
\pgfsetlinewidth{1.204500pt}%
\definecolor{currentstroke}{rgb}{0.368627,0.337255,0.870588}%
\pgfsetstrokecolor{currentstroke}%
\pgfsetdash{}{0pt}%
\pgfpathmoveto{\pgfqpoint{0.806557in}{0.060000in}}%
\pgfpathlineto{\pgfqpoint{1.006557in}{0.060000in}}%
\pgfusepath{stroke}%
\end{pgfscope}%
\begin{pgfscope}%
\definecolor{textcolor}{rgb}{0.000000,0.000000,0.000000}%
\pgfsetstrokecolor{textcolor}%
\pgfsetfillcolor{textcolor}%
\pgftext[x=1.051557in,y=0.035000in,left,base]{\color{textcolor}{\rmfamily\fontsize{8.000000}{9.600000}\selectfont Rprop}}%
\end{pgfscope}%
\begin{pgfscope}%
\pgfsetbuttcap%
\pgfsetroundjoin%
\pgfsetlinewidth{1.204500pt}%
\definecolor{currentstroke}{rgb}{0.800000,0.447059,0.000000}%
\pgfsetstrokecolor{currentstroke}%
\pgfsetdash{}{0pt}%
\pgfpathmoveto{\pgfqpoint{1.508252in}{0.060000in}}%
\pgfpathlineto{\pgfqpoint{1.708252in}{0.060000in}}%
\pgfusepath{stroke}%
\end{pgfscope}%
\begin{pgfscope}%
\definecolor{textcolor}{rgb}{0.000000,0.000000,0.000000}%
\pgfsetstrokecolor{textcolor}%
\pgfsetfillcolor{textcolor}%
\pgftext[x=1.753252in,y=0.035000in,left,base]{\color{textcolor}{\rmfamily\fontsize{8.000000}{9.600000}\selectfont CR-FM-NES}}%
\end{pgfscope}%
\begin{pgfscope}%
\pgfsetbuttcap%
\pgfsetroundjoin%
\pgfsetlinewidth{1.204500pt}%
\definecolor{currentstroke}{rgb}{0.298039,0.352941,0.015686}%
\pgfsetstrokecolor{currentstroke}%
\pgfsetdash{}{0pt}%
\pgfpathmoveto{\pgfqpoint{2.515486in}{0.060000in}}%
\pgfpathlineto{\pgfqpoint{2.715486in}{0.060000in}}%
\pgfusepath{stroke}%
\end{pgfscope}%
\begin{pgfscope}%
\definecolor{textcolor}{rgb}{0.000000,0.000000,0.000000}%
\pgfsetstrokecolor{textcolor}%
\pgfsetfillcolor{textcolor}%
\pgftext[x=2.760486in,y=0.035000in,left,base]{\color{textcolor}{\rmfamily\fontsize{8.000000}{9.600000}\selectfont MR15-GA}}%
\end{pgfscope}%
\begin{pgfscope}%
\pgfsetbuttcap%
\pgfsetroundjoin%
\pgfsetlinewidth{1.204500pt}%
\definecolor{currentstroke}{rgb}{0.000000,0.000000,0.000000}%
\pgfsetstrokecolor{currentstroke}%
\pgfsetdash{}{0pt}%
\pgfpathmoveto{\pgfqpoint{3.422078in}{0.060000in}}%
\pgfpathlineto{\pgfqpoint{3.622078in}{0.060000in}}%
\pgfusepath{stroke}%
\end{pgfscope}%
\begin{pgfscope}%
\definecolor{textcolor}{rgb}{0.000000,0.000000,0.000000}%
\pgfsetstrokecolor{textcolor}%
\pgfsetfillcolor{textcolor}%
\pgftext[x=3.667078in,y=0.035000in,left,base]{\color{textcolor}{\rmfamily\fontsize{8.000000}{9.600000}\selectfont RL2CO}}%
\end{pgfscope}%
\begin{pgfscope}%
\pgfsetbuttcap%
\pgfsetroundjoin%
\pgfsetlinewidth{1.204500pt}%
\definecolor{currentstroke}{rgb}{0.694118,0.125490,0.513725}%
\pgfsetstrokecolor{currentstroke}%
\pgfsetdash{{4.440000pt}{1.920000pt}}{0pt}%
\pgfpathmoveto{\pgfqpoint{4.197211in}{0.060000in}}%
\pgfpathlineto{\pgfqpoint{4.397211in}{0.060000in}}%
\pgfusepath{stroke}%
\end{pgfscope}%
\begin{pgfscope}%
\definecolor{textcolor}{rgb}{0.000000,0.000000,0.000000}%
\pgfsetstrokecolor{textcolor}%
\pgfsetfillcolor{textcolor}%
\pgftext[x=4.442211in,y=0.035000in,left,base]{\color{textcolor}{\rmfamily\fontsize{8.000000}{9.600000}\selectfont Random schedule}}%
\end{pgfscope}%
\begin{pgfscope}%
\pgfsetbuttcap%
\pgfsetroundjoin%
\pgfsetlinewidth{1.204500pt}%
\definecolor{currentstroke}{rgb}{0.533333,0.533333,0.533333}%
\pgfsetstrokecolor{currentstroke}%
\pgfsetdash{{1.200000pt}{1.980000pt}}{0pt}%
\pgfpathmoveto{\pgfqpoint{5.414045in}{0.060000in}}%
\pgfpathlineto{\pgfqpoint{5.614045in}{0.060000in}}%
\pgfusepath{stroke}%
\end{pgfscope}%
\begin{pgfscope}%
\definecolor{textcolor}{rgb}{0.000000,0.000000,0.000000}%
\pgfsetstrokecolor{textcolor}%
\pgfsetfillcolor{textcolor}%
\pgftext[x=5.659045in,y=0.035000in,left,base]{\color{textcolor}{\rmfamily\fontsize{8.000000}{9.600000}\selectfont Random search}}%
\end{pgfscope}%
\begin{pgfscope}%
\pgfsetbuttcap%
\pgfsetroundjoin%
\pgfsetlinewidth{1.204500pt}%
\definecolor{currentstroke}{rgb}{0.000000,0.000000,0.000000}%
\pgfsetstrokecolor{currentstroke}%
\pgfsetdash{{1.200000pt}{1.980000pt}}{0pt}%
\pgfpathmoveto{\pgfqpoint{6.544977in}{0.060000in}}%
\pgfpathlineto{\pgfqpoint{6.744977in}{0.060000in}}%
\pgfusepath{stroke}%
\end{pgfscope}%
\begin{pgfscope}%
\definecolor{textcolor}{rgb}{0.000000,0.000000,0.000000}%
\pgfsetstrokecolor{textcolor}%
\pgfsetfillcolor{textcolor}%
\pgftext[x=6.789977in,y=0.035000in,left,base]{\color{textcolor}{\rmfamily\fontsize{8.000000}{9.600000}\selectfont VBS}}%
\end{pgfscope}%
\end{pgfpicture}%
\makeatother%
\endgroup%
\\
  \gfitem{media/ertd/rtd_bbob_headroom_train.pgf}{Training set}{fig:rtd_bbob_headroom_train}
  \gfitem{media/ertd/rtd_group_dimensionality_2.pgf}{2-D}{fig:rtd_dim_2}
  \gfitem{media/ertd/rtd_group_dimensionality_3.pgf}{3-D}{}
  \gfitem{media/ertd/rtd_group_dimensionality_5.pgf}{5-D}{}
  \gfitem{media/ertd/rtd_group_dimensionality_10.pgf}{10-D}{}
  \gfitem{media/ertd/rtd_group_dimensionality_20.pgf}{20-D}{}
  \gfitem{media/ertd/rtd_group_dimensionality_40.pgf}{40-D}{}
  \gfitem{media/ertd/rtd_group_dimensionality_64.pgf}{64-D}{}
  \gfitem{media/ertd/rtd_group_dimensionality_256.pgf}{256-D}{}
  \gfitem{media/ertd/rtd_group_dimensionality_1024.pgf}{1024-D}{fig:rtd_dim_1024}
  \gfitem{media/ertd/rtd_group_noise_False.pgf}{Deterministic}{fig:rtd_noise_deterministic}
  \gfitem{media/ertd/rtd_group_noise_True.pgf}{Stochastic}{fig:rtd_noise_stochastic}
  \caption{Empirical runtime distributions over the benchmark function problems. (a) the training set as a whole; (b)--(j) the test set grouped by problem dimensionality $d$; and (k)--(l) the same test set split by deterministic or stochastic tasks.}
  \label{fig:rtd_groups}
\end{gridfigure}

\section{Additional Evaluation Families}
\label{app:families}

\subsection{Random Quadratics}
\label{app:families-quadratics}

\noindent Each quadratic problem instance is defined as follows: we draw $\boldsymbol{W} \in \mathbb{R}^{d \times d}$ and $\boldsymbol{y} \in \mathbb{R}^{d}$ with independent standard normal entries, and define
\[
f(\boldsymbol{x}) = \lVert \boldsymbol{W}\boldsymbol{x} - \boldsymbol{y} \rVert^{2}.
\]
The Hessian is $2\boldsymbol{W}^{\top}\boldsymbol{W}$, so the conditioning of each instance depends entirely on its random draw. A square matrix with independent Gaussian entries is nonsingular with probability one, so $\boldsymbol{W}\boldsymbol{x} = \boldsymbol{y}$ admits a unique solution and the global minimum is zero. The benchmark has 90 instances, each with its own random seed and dimension $d$ uniformly sampled from 10 to 50.

\subsection{Gaussian-Blob Classification}
\label{app:families-gaussian}

\noindent Each instance trains a small classifier on a synthetic two-dimensional dataset, following the neural-network benchmark of \textcite{Li2017}. The data consist of 100 points from four multivariate Gaussian components, each contributing 25 points. For component $i$, the mean is $5\boldsymbol{\mu}_{i}$ with $\boldsymbol{\mu}_{i} \sim \mathcal{N}(\boldsymbol{0}, \boldsymbol{I})$, and the covariance is $\boldsymbol{A}_{i}\boldsymbol{A}_{i}^{\top}$, where $\boldsymbol{A}_{i}$ is drawn uniformly from $[0,1]^{2 \times 2}$. Each component is assigned one of two labels, with both classes present. The classifier is a multilayer perceptron (MLP) with two inputs, ReLU hidden units, and a two-way softmax output. Training uses full-batch cross-entropy loss with $L_{2}$ regularization of $5 \times 10^{-4}$ on every weight.

The family independently adjusts the dataset and model, combining 11 random seeds with depths of 2, 3, or 4 hidden layers and widths of 2, 4, or 6 units, yielding problems with 18--158 trainable parameters. For each instance, the global minimum is approximated via three optimization restarts. Each restart keeps the better result from a 1000-step cosine-annealed Adam run and a 300-step L-BFGS run. The first restart uses the original initialization; subsequent restarts use resampled weights. The smallest loss over all restarts defines the reference baseline for all targets.

\subsection{Meta-Optimization of Adam}
\label{app:families-meta}

\noindent The lower-level problem is a classification instance from Appendix~\ref{app:families-gaussian}, using the smallest architecture in its grid: an MLP with two inputs, two hidden layers of width two, and two outputs, trained on a similarly generated dataset. The outer optimization tunes three Adam hyperparameters in $[0,1]^3$: the learning rate $10^{-5 + 7x_{1}}$ and momentum terms $\beta_{1} = 0.85 + 0.15x_{2}$ and $\beta_{2} = 0.85 + 0.15x_{3}$.

To evaluate the objective at $\boldsymbol{x} = (x_1, x_2, x_3)$, we train an initialized MLP with Adam for 20 steps on the lower-level loss: softmax cross-entropy with $L_{2}$ regularization $5 \times 10^{-4}$. One outer evaluation corresponds to a full inner training run, and the outer gradient is obtained by backpropagating through all unrolled steps. With fixed initialization per instance, the objective is deterministic but its global minimum is not analytically tractable. We approximate it using Adam with five independent restarts of 300 outer steps, taking the lowest loss as reference. All targets are defined relative to this empirical best, and the problem family has ten instances differing only by random seed.

\printbibliography

\end{document}